\documentclass{article}
\usepackage[utf8]{inputenc}
\usepackage{mathtools}
\usepackage{authblk}
\usepackage{amsmath}
\usepackage{amssymb}
\usepackage{bigints}
\usepackage{graphicx}
\usepackage{gensymb}
\usepackage{float}
\usepackage[normalem]{ulem}
\usepackage{comment}
\usepackage{subfig}
\usepackage{geometry}
\usepackage{booktabs}
\usepackage{hyperref}
\usepackage{xurl}
\usepackage{xcolor}
\begin{document}

\title{Analyzing Spatio-Temporal Dynamics of Dissolved Oxygen for the River Thames using Superstatistical Methods and Machine Learning}

\author{Hankun He$\,^{1*}$, Takuya Boehringer$\,^2$, Benjamin Schäfer$\,^3$, Kate Heppell$\,^{4,5}$, Christian Beck}

\affil[1]{Centre for Complex Systems, Queen Mary University of London, UK}
\affil[2]{University College London, UK}
\affil[3]{Institute for Automation and Applied Informatics, Karlsruhe Institute of Technology, Germany}
\affil[4]{Chilterns National Landscape, UK}
\affil[5]{School of Geography, Queen Mary University of London, UK}



\maketitle

\begin{abstract}
By employing superstatistical methods and machine learning, we analyze time series data of water quality indicators for the River Thames (UK). The indicators analyzed include dissolved oxygen, temperature, electrical conductivity, pH, ammonium, turbidity, and rainfall, with a specific focus on the dynamics of dissolved oxygen. After detrending, the probability density functions of dissolved oxygen fluctuations exhibit heavy tails that are effectively modeled using $q$-Gaussian distributions. Our findings indicate that the multiplicative Empirical Mode Decomposition method stands out as the most effective detrending technique, yielding the highest log-likelihood in nearly all fittings. We also observe that the optimally fitted width parameter of the $q$-Gaussian shows a negative correlation with the distance to the sea, highlighting the influence of geographical factors on water quality dynamics. In the context of same-time prediction of dissolved oxygen, regression analysis incorporating various water quality indicators and temporal features identify the Light Gradient Boosting Machine as the best model. SHapley Additive exPlanations reveal that temperature, pH, and time of year play crucial roles in the predictions. Furthermore, we use the Transformer, a state-of-the-art machine learning model, to forecast dissolved oxygen concentrations. For long-term forecasting, the Informer model consistently delivers superior performance, achieving the lowest Mean Absolute Error and Symmetric Mean Absolute Percentage Error with the 192 historical time steps that we used. This performance is attributed to the Informer's ProbSparse self-attention mechanism, which allows it to capture long-range dependencies in time-series data more effectively than other machine learning models. It effectively recognizes the half-life cycle of dissolved oxygen, with particular attention to critical periods such as morning to early afternoon, late evening to early morning, and key intervals between the 16th and 26th quarter-hours of the previous half-day. Our findings provide valuable insights for policymakers involved in ecological health assessments, aiding in accurate predictions of river water quality and the maintenance of healthy aquatic ecosystems.

\end{abstract}

\section*{Introduction}
The River Thames (UK) plays a pivotal role in sustaining the city of London, supplying most of the drinking water for its residents. However, the ongoing sewage crisis has sparked deep concerns over its potential effects on the ecosystem (endangered aquatic life), human health (contamination of drinking water), and the economy (decreased recreational activities). When millions of tonnes of sewage enter the Thames, the water quality dynamics are profoundly disturbed. For example, the levels of dissolved oxygen (DO), a key water quality indicator essential for many aquatic organisms and reflective of river metabolic pulses, could drop substantially. Of particular interest are extreme event situations. If DO falls below a certain threshold level, which differs for different organisms, the consequences can be disastrous for aquatic life~\cite{mccormick2021state}. Thus, maintaining healthy water conditions in the River Thames has become a pressing and escalating challenge. As researchers, we aim to shed light on the spatio-temporal complex dynamics of water quality, specifically DO, to potentially assist policymakers and environmental agencies in effectively managing river conditions and maintaining healthy aquatic environments. The findings in our paper, to be presented in detail in the following sections, could provide tools for accurately forecasting DO concentrations, which would help alert stakeholders when forecasted DO values fall into ranges that endanger aquatic life. Our findings could also help to identify critical time windows where interventions, such as oxygen injection or controlled water treatment releases, should be implemented, alongside enhanced monitoring at sites with more extreme variations. Future studies may extend our methodology to other observables, such as electrical conductivity, in different river systems also outside the UK. 
Our approach in the following is based on two main pillars: superstatistics and machine learning.

Superstatistical methods, originating from turbulence research \cite{beck2007statistics,PhysRevE.72.056133,BECK2003267}, offer a powerful effective approach to describe the dynamics of complex nonequilibrium systems, assuming the existence of well-separated time scales. These methods have been applied to a wide range of fields, such as high energy scattering processes~\cite{beck2009superstatistics,Sevilla2019,Ayala2020}, Ising systems~\cite{Cheraghalizadeh2021}, cosmic ray physics~\cite{Yalcin2018,Smolla_2020}, plasma physics~\cite{livadiotis2017kappa,Davis2019}, ultracold gases~\cite{PhysRevLett.118.143401}, non-Gaussian diffusion processes in small complex systems~\cite{PhysRevX.7.021002,Itto2021}, power grid frequency~\cite{schafer2018non}, bacterial DNA~\cite{Bogachev20171}, financial time series~\cite{GIDEA2018820,UCHIYAMA2019120930} and traffic delays~\cite{BRIGGS2007498,mitsokapas2021statistical}. Recent years have also witnessed an increase in its application in environmental sciences, including wind statistics~\cite{Weber20191}, air pollution statistics~\cite{Williams2020,he2022spatial}, rainfall statistics~\cite{de2018superstatistical}, sea surface currents~\cite{npg-30-515-2023}, earthquake swarms~\cite{Sardeli_2023,agarap2018deep} and water quality time series~\cite{SCHAFER2021102881}. The dynamics of river water quality indicators, which involve a complex interplay of physical, chemical, and biological processes, can be viewed as a superposition of various dynamics across different time scales. These include day-night cycles, seasonal temperature fluctuations, long-term trends due to global warming and pollution events, making this a suitable subject for the superstatistical approach which requires complex dynamics on well-separated time scales. Water quality time series (such as DO concentrations and electrical conductivity) as measured in the River Chess have been previously shown to exhibit superstatistical features~\cite{SCHAFER2021102881}. However, the extent to which the water quality dynamics in other, much bigger rivers can be explained using similar techniques based on superstatistics remains unclear so far, as do the environmental insights and forecasts that can be systematically derived from the superstatistical approach. In this paper we will fill this gap and present a systematic analysis for new data from the tidal section of the River Thames.

Machine learning, as a complementary approach to statistical analysis, has been extensively utilized to predict water quality time series in various rivers, aiding in the understanding of the fluctuating dynamics of water quality. For example, Generalized Additive Models and gradient boosting methods have been employed to predict electrical conductivity in the River Chess~\cite{schafer2022machine}. In the Sakarya Basin in Turkey, models such as deep learning, support vector machine regression (SVMR), Gaussian process regression, and artificial neural networks (ANNs) were used to estimate dissolved oxygen levels, with ANNs proving to be the most effective~\cite{10.1007/978-3-031-43169-2_21}. ANNs were also shown to be the most effective to predict suspended sediment concentration in the Thames River, Ontario~\cite{mohamed2018suspended}. Additionally, SVMR, decision tree regression and extra tree regression were used to predict the water quality index (WQI) of the Lam Tsuen River in Hong Kong~\cite{ASADOLLAH2021104599}. A Long short term memory (LSTM) model has been effective in predicting DO levels across the watersheds across the U.S.~\cite{doi:10.1021/acs.est.0c06783}. In the context of the River Thames, existing literature includes neural network applications for chlorophyll estimation and Random Forest methods for predicting nutrient concentrations in two tributaries of the River Thames~\cite{castrillo2020estimation,SAINZPARDODIAZ2023120726}, applications of Random Forest and Gradient Boosting models in estimating nitrate concentrations in surface waters~\cite{https://doi.org/10.1002/lom3.10468}, and neural network applications to river flow forecasting~\cite{zounemat2021comprehensive}. The state-of-the-art technique, Transformer, driven by the attention mechanism~\cite{vaswani2017attention}, excels in sequence forecasting but has been rarely utilized in the context of water quality indicator dynamics. While the study~\cite{nair2023temporal} applied a Transformer model to forecast the river WQI, which combines various water quality parameters to provide an overall assessment of water quality, it differs from predicting individual indicators, which offers detailed insights into specific indicator dynamics, thereby enabling targeted investigations.

Our research, to be described in detail in the following sections, innovates in the analysis and forecasting of spatio-temporal DO dynamics within river systems. We introduce a novel multiplicative detrending method for the tidal River Thames, which proves to be more effective in handling complex tidal variations. After detrending, the superstatistical analysis reveals that DO fluctuations exhibit $q$-Gaussian distributions with power-law tails. By systematically extracting the optimum fitted superstatistical parameters, we uncover important correlations between these parameters and geographical factors, particularly proximity to the sea—this represents a novel insight. Our descriptive statistical parameters are extracted from the analyzed time series using the standard methods of superstatistics~\cite{beck2004superstatistics}. In terms of machine learning, we demonstrate the robustness of the Light Gradient Boosting Machine (LGBM) model for predicting DO and used SHAP values to identify temperature, pH, and time of year as dominant predictors of DO fluctuations. We use the Informer model, a novel Transformer architecture that has been rarely applied in the study of water quality dynamics. Our results demonstrate the Informer’s superior performance, achieving the lowest errors for long-term forecasts. Moreover, the ProbSparse self-attention mechanism of the Informer model dynamically focuses on the most informative parts of a long input sequence, inherently capturing long-range dependencies and temporal dynamics. This capability addresses areas where classical machine learning models often face limitations without extensive data preprocessing. For a detailed explanation, see the 'Forecasting models' subsection. When forecasting 48 quarter-hour time steps of DO, the model identifies crucial time windows in the input data, specifically during morning to early afternoon and late evening to early morning of the most recent time steps, as well as between the 16th and 26th quarter-hour of the preceding half-day, which contributed to its superior performance. This identification provides novel insights into the life cycles of DO dynamics. Overall, our research not only advances the theoretical understanding of DO dynamics in diverse river settings but also enhances practical capabilities in monitoring and forecasting water quality, which is crucial for environmental management and policy-making.

Key questions include striving for a deeper understanding of the complex system interactions and processes relevant for the River Thames: Can DO concentrations in the river be explained via superstatistics, and can we comprehensively account for extreme DO events arising from environmental processes and human activities? How do different water quality indicators affect the prediction of DO? How can we construct a powerful approach for effectively forecasting water quality indicators? These important types of questions will be dealt with in the following, with data from the different measuring stations positioned along the tidal stretch.

This paper is structured as follows: First, we describe our data sets from nine water quality monitoring sites along the River Thames. We then employ statistical and detrending methods to isolate the DO fluctuations from the trend. Next, we apply a superstatistical approach to these fluctuations and uncover the relevant superstatistical parameters as a function of the sites' distances to the sea. Following this, we utilize regression-based machine learning techniques to predict same-time DO levels, highlighting an interpretation of the model's prediction mechanism. This sheds light on the most important features influencing the prediction. We then employ state-of-the-art machine learning methodologies, including Informer as a Transformer variant, and deduce the method with the best forecasting performance for DO. Lastly, we explore the self-attention mechanism underlying the Informer model, emphasizing attention weights to explain which inputs contribute to its superior performance.

\section*{The data available}
In this paper, we utilise data from nine monitoring sites along a $45$ km stretch of the River Thames, sourced from Meteor Communications' telemetry system~\cite{Meteor}. This system connects to multi-parameter water quality sondes within a sensor web network across the River Thames, jointly managed by the Environment Agency(EA) and Meteor Communications. The sites span the Upper Tideway, Central Tideway, and Estuary regions of the Thames. This stretch marks the tidal portion of the river, which is significant for understanding the complex interactions and variability caused by tidal movements, which are relevant for ecosystems and for pollution concentrations alike. This segment of the river also passes through central London.

\begin{figure}[h!]
    \centering
    \includegraphics[width=1\textwidth]{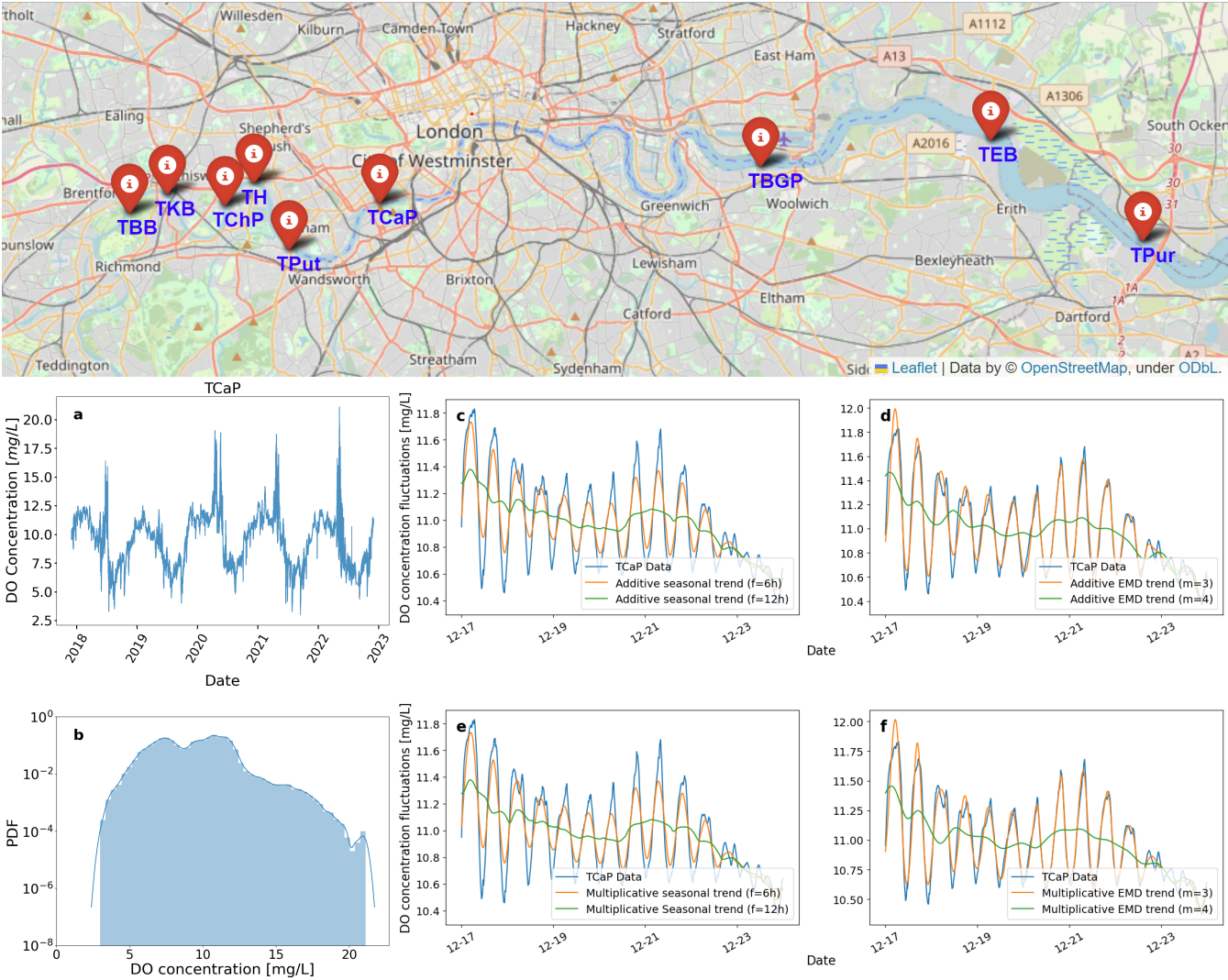}
    \caption{Top: A map of the nine available water quality monitoring sites (red markers) along the River Thames. The geographical visualisations were generated using the Folium Python library~\cite{Folium} with map data sourced from OpenStreetMap~\cite{OpenStreetMap}. We show the trajectory (a) and PDF (b) of the DO concentration at TCaP over a time span of five years. We apply additive seasonal (c), additive EMD (d), multiplicative seasonal (e) and multiplicative EMD (f) detrending methods. A filtering frequency of $f = 6$ hours or dropping $m = 3$ modes (orange) captures the oscillating trend while preserving the short-term fluctuations.}
    \label{fig:intro}
\end{figure}

To provide a geographical overview of our analysed data, we display all the data sites’ locations on a map in Fig.~\ref{fig:intro}. The nine monitoring sites stretch from West to East along the River Thames as follows: Thames Brentford Barge (TBB) and Thames Kew Barge (TKB) in the Upper Tideway region, Thames Chiswick Pier (TChP), Thames Hammersmith (TH), Thames Putney (TPut), Thames Cadogan Pier (TCaP), Thames Barrier Gardens Pier (TBGP), Thames Erith Barge (TEB), and Thames Purfleet (TPur) in the Central Tideway and Estuary regions. The time series measured span from 01/12/2017 to 01/12/2022 and have a temporal resolution of $15$ minutes, except for the rainfall feature. To maintain the integrity of all nine data sets, each underwent some checks and data processing prior to being incorporated into the analysis. Details of the data processing methods are provided in the 'Methods' section.

Examples of DO concentration time series from site TCaP are shown in Fig.~\ref{fig:intro}a. We observe strong seasonality in the trajectory, with summer exhibiting lower DO levels as compared to the winter. This can be attributed to the temperature-dependent solubility of DO, as cold water holds more DO than warm water does. Site TCaP also experiences seasonal spikes during the early summer season. This phenomenon likely reflects a period of algal blooms that is fueled by increased sunlight. This results in elevated photosynthetic activity, leading to an increase in DO levels. In Fig.~\ref{fig:intro}b, the probability density function (PDF) of DO shows clear deviations from Gaussianity, in agreement with previous results found for smaller rivers~\cite{SCHAFER2021102881}. A large portion of the observed variability arises not only from seasonal cycles but also from daily cycles, which must be accounted for before we apply our statistical analysis. In the following section, we will describe how to separate the fluctuations from the trend.

\section*{Detrending}

Rather than modeling the complete distribution, including its daily and seasonal variations, our approach here is centered on describing the fluctuations around the mean. The mean can be time-dependent and is denoted as 'trend' in the following. To initiate the detrending procedure, 
we may decompose the measured time series in an additive way as
\begin{equation}
y_t = T_t + F_t+ R_t.
\label{eq:add}
\end{equation}
Here $y_t$ is the time series, $T_t$ is the trend, $F_t$ are the fluctuations evolving on a much faster time scale, and $R_t$ is the remainder, all evaluated at time $t$. 
Alternatively, one may also consider a multiplicative model, in which case
we decompose the time series as
\begin{equation}
y_t = \hat{T}_t\times \hat{F}_t \times \hat{R}_t.
\label{eq:mult}
\end{equation}
To effectively retrieve the fluctuations, we employ seasonal decomposition and empirical mode decomposition (EMD). Seasonal decomposition separates the trend within the data by applying a moving average with a specific filtering frequency, denoted as $f$. The difference between this moving average and the original data is characterised as fluctuation. Alternatively, the EMD decomposes the full trajectory into its ordered intrinsic mode functions (IMFs) ranging from slowly changing to highly oscillating modes as follows:
\begin{equation}
y_t = \sum_{i=1}^{N}{IMF_i(t)+r_N(t)},
\label{eq:emd}
\end{equation}
where $N$ is the number of modes, and $r_N(t)$ represents the residue corresponding to N modes. The trend component is identified by summing over the first few IMFs:
\begin{equation}
T_t = \sum_{i=1}^{N-m}{IMF_i(t)},
\label{eq:emd_trend}
\end{equation}
where the decision to omit $m$ modes to define the trend is somewhat arbitrary and has to be made by the user. After removing the IMFs that constitute the trend, we consider the remaining detrended data as representing the fluctuations:
\begin{equation}
F_t = \sum_{i=N-m+1}^{N}{IMF_i(t)+r_N(t)}.
\label{eq:emd_fluc}
\end{equation}
Given the additive nature of EMD, when applying a multiplicative decomposition, a logarithmic transformation is applied to Eq.~(\ref{eq:mult}) to obtain
\begin{equation}
log(y_t) = log(\hat{T}_t) + log(\hat{F}_t)+ log(\hat{R}_t)
\label{eq:emd_log}
\end{equation}
which reduces the problem to an additive one. For this we have to assume that all
functions $\hat{T}_t,\hat{F}_t,\hat{R}_t$ are positive so that the logarithm is well-defined.

\begin{figure}[h]
    \centering
    \subfloat{\includegraphics[width=0.65\textwidth]{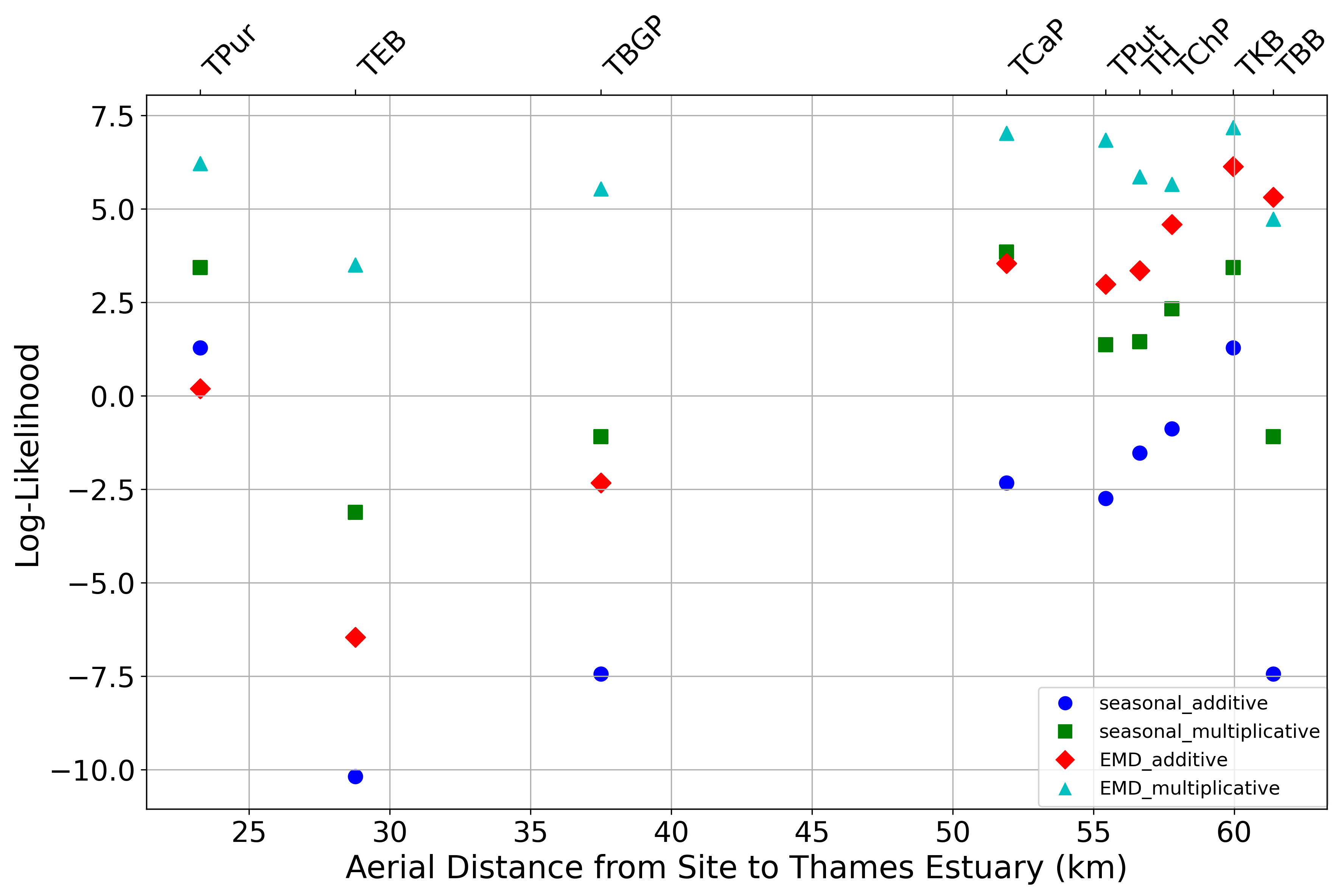}}
    \caption{Scatter plot showing the relationship between log-likelihood values for different detrending methods and the distance to sea for various sites. The log-likelihood values are evaluated based on the best fit to a $q$-Gaussian distribution. The graph demonstrates the superiority of the multiplicative methods over the additive ones, with the multiplicative EMD method standing out as the most effective approach in nearly all cases.}
    \label{fig:log}
\end{figure}

Which detrending decomposition method is most suitable in the context of water quality dynamics? We wish to have a simple description of the fluctutions in terms of a simple stochastic process that has a $q$-Gaussian probability density (see next section for more details on the definition of $q$-Gaussians in the superstatistical context). The relevance of $q$-Gaussians was noticed in previous work for data from the River Chess
\cite{schafer2022machine}. Hence, for each measuring station, we computed the log-likelihood for a best fit of the observed PDF of the fluctuations by a $q$-Gaussian probability distribution. The result is shown in Fig.~\ref{fig:log} for the different distances to the sea and detrending methods. One immediately noticeable observation is that, regardless of whether the chosen method is seasonal or EMD, the multiplicative method generally produces a slightly better fit with $q$-Gaussians than the additive decomposition, as indicated by higher log-likelihood values. Additionally, coastal conditions are more challenging for the additive method, while the multiplicative methods remain relatively consistent regardless of the distance to the sea. Ultimately, the multiplicative EMD method appears to be the most effective detrending method, exhibiting the highest log-likelihoods with one exception only. Our approach advances beyond the additive methods previously applied to the River Chess~\cite{schafer2022machine}. We introduce a novel multiplicative EMD method to address the pronounced tidal periodic variations observed in the DO time series relevant for the River Thames. Our results confirm that this approach effectively extracts $q$-Gaussian fluctuation spectra, demonstrating its suitability for complex DO time series that vary significantly in amplitude and frequency.

Note that the detrended data still include some remaining level of noise, which we choose not to remove. The remainder component, or noise, usually refers to random, unexplained variations in the data that are not part of the trend or of the fluctuations. One might argue that removing the remainder from fluctuations would yield clearer insights into the inherent dynamics of DO. However, our analysis aims to understand the behavior of realistic data, including natural variability and extreme events. Hence, we do not isolate unexplained noise from the detrended data, considering both as integral parts of the overall fluctuations in DO.

The results of different detrending methods, using as an example DO levels measured at the site TCaP over a one-week period, are illustrated in Fig.~\ref{fig:intro}(c-f). For the seasonal decomposition (c,e), we choose a filter frequency of $f=6$ hours, and for EMD (d,f), we drop $m=3$ modes. The filter frequency of $f=12$ hours (green curves) aligns with the periodicity of DO, hence would provide a nearly constant trend function without oscillations. However, with the choice of $f=6$ hours, the trend (orange curve) reflects more the reality of periodic average variations in the data. For this choice we get a meaningful description of fluctuations on a short time scale that are added to the oscillating trend. Similarly, for EMD, we opt for dropping $m=3$ modes (orange curves) over the smoother trend (green curves) to extract the short timescale fluctuations. With the data now detrended, we continue to explore the fluctuation statistics using a superstatistical approach.

\section*{Superstatistical analysis}
The DO fluctuations are the result of the complex interplay of physical, chemical, and biological processes in the River Thames's ecosystem, and are within the realm where superstatistical approaches are suitable for an effective description. The core concept of superstatistics~\cite{PhysRevE.72.056133,BECK2003267} is that a complex, heavy-tailed probability distribution describing the statistics of a long time series is produced by a superposition of PDFs of many shorter time series, each with a non-heavy-tailed distribution. As shown previously in~\cite{SCHAFER2021102881}, the marginal DO fluctuation distribution can indeed be seen as an aggregation of many simple Gaussian distributions of fluctuating variance, leading us to expect that Tsallis $q$-Gaussian distributions are effectively relevant for this problem. The $q$-Gaussian distribution~\cite{tsallis1988possible} is a probability distribution that generalises the standard Gaussian distribution. It maximizes Tsallis entropy subject to suitable constraints. The $q$-Gaussian distribution emerges naturally within the superstatistical framework, particularly in systems that exhibit fluctuations in an inverse variance parameter $\hat{\beta}$
on a larger time scale. $\hat{\beta}$ can in principle follow any kind of distribution.
If the PDF of $\hat{\beta}$ is sharply peaked, then this in good approximation leads to a $q$-Gaussian distribution when integrating out the $\hat{\beta}$ fluctuations~\cite{beck2001dynamical,PhysRevE.72.056133}.
This statement is exact if  $\hat{\beta}$ follows a $\chi^2$-distribution, as described by
\begin{equation}
f(\hat{\beta}) = \frac{\hat{\beta}^{\frac{n}{2} - 1}}{\Gamma(\frac{n}{2})} \left(\frac{n}{2\hat{\beta}}^{\frac{n}{2}}\right)  e^{-\frac{n\hat{\beta}}{2\hat{\beta}_0}},
\label{chi_sq}
\end{equation}
where $n$ denotes the degrees of freedom and $\hat{\beta}_0$ represents the mean of $\hat{\beta}$. In this case the marginal distribution 
\begin{equation}
p(x)=\int_0^\infty f(\hat{\beta}) p(x|\hat{\beta}) d\hat{\beta}
\end{equation}
is exactly a $q$-Gaussian if $p(x|\hat{\beta)}$ is an ordinary Gaussian with
variance parameter $\hat{\beta}$. 
The $q$-Gaussian PDF, which is well motivated by superstatistics, is given by the formula
\begin{equation}
p(x) = \frac{1}{C_q} [ 1 + (q-1) \beta (x - \mu)^2 ]^{\frac{1}{1-q}},
\label{q_gaus}
\end{equation}
where $C_q$ is a normalisation constant, $\mu$ is a shift parameter, $q$ is a shape parameter, also known as the entropic index, and $\beta$ is a scale parameter that is proportional to the expectation $\langle \hat{\beta} \rangle$. The shape parameter $1<q<3$ controls the degree of non-extensivity of the system and the distribution reduces to the Gaussian distribution when $q=1$. Different from the standard Gaussian distribution, the $q$-Gaussian distribution depends on two parameters, the scale parameter $\beta$ and the shape parameter $q$, making it quite an effective tool for fitting power-law tails describing extreme events. 

\begin{figure}[h]
    \centering
    \includegraphics[width=1\textwidth]{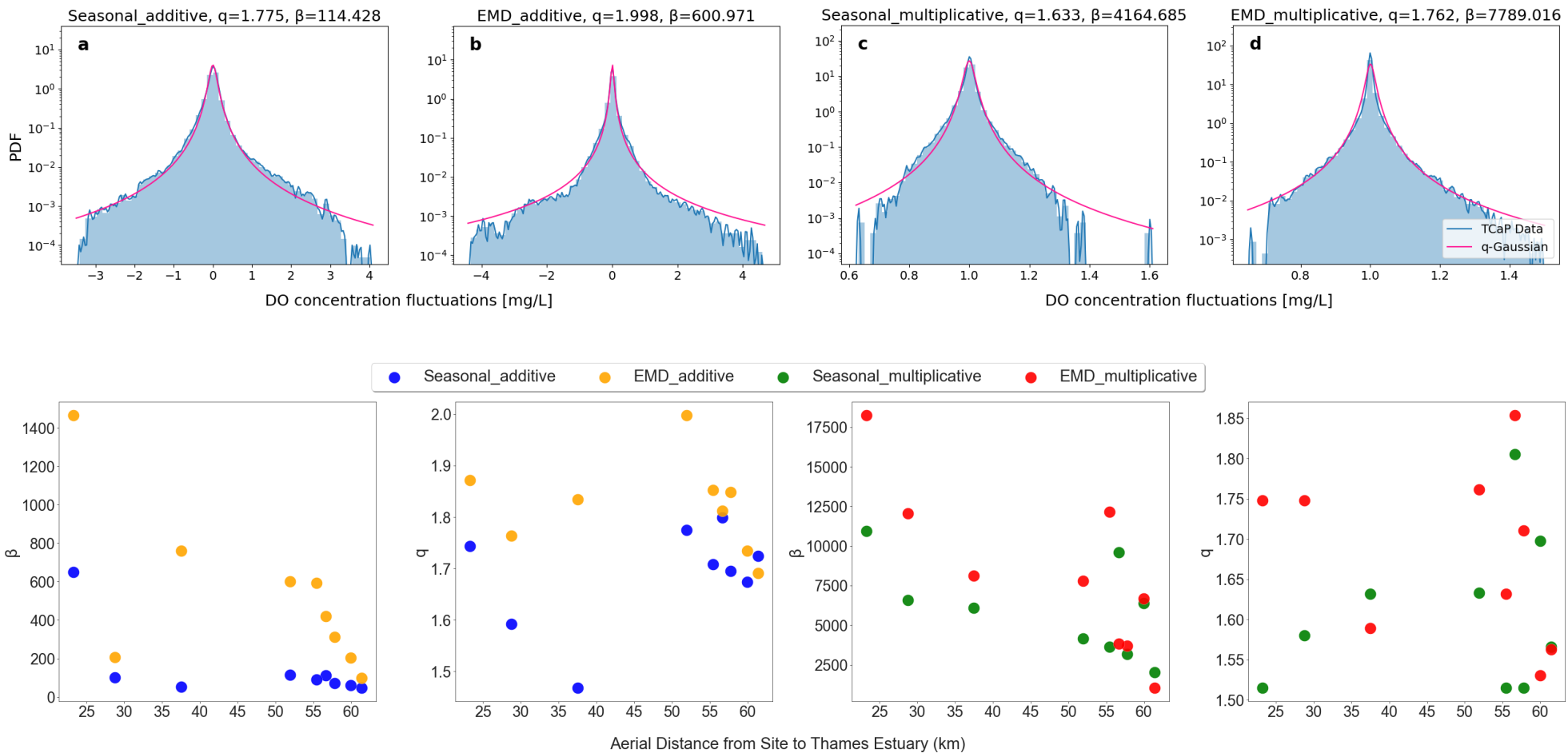}
    \caption{Top: PDFs of oxygen fluctuations obtained via additive seasonal (a), additive EMD (b), multiplicative seasonal (c) and multiplicative EMD (d) detrending methods, for the example of the site TCaP. Regardless of the methods used, detrending leads to non-Gaussian distributions, which can be approximated by $q$-Gaussian distributions (purple). 
    Bottom: For each site along the River Thames, we plot the scale parameter $\beta$ and the shape parameter $q$ from the $q$-Gaussian fitting against its distance to the Thames Estuary/sea.}
    \label{fig:beta_q}
\end{figure}

The measured DO fluctuation distributions (blue) follow a $q$-Gaussian distribution (purple) in good approximation, as illustrated in Fig.~\ref{fig:beta_q}(a-d) for the example of the site TCap. The PDFs are well approximated by the $q$-Gaussian fits regardless of which detrending method is applied. Note that these are the distributions of the deviation of the DO signal from the systematic trend, not the signal itself. We optimized the fitting parameters via maximum likelihood estimation (MLE) for all nine sites.

By employing the optimal $q$-Gaussian fittings, we determine the scale parameter $\beta$ and the shape parameter $q$ for each site. Recall that $\beta$ adjusts the width of the distribution whereas $q$ determines the tail behavior of the distribution. To identify underlying spatial patterns in the variability of the relevant parameters, we plot each of these parameters against their respective distances to the Thames Estuary in Fig.~\ref{fig:beta_q}(bottom). The first general observation is the non-Gaussianity of DO fluctuations for all nine measuring sites, as indicated by $q>1$. We confirm that the distributions of DO fluctuations exhibits a heavy-tailed characteristic in the more complex environment of the River Thames, compared to the relatively smaller River Chess. The majority of these $q$ values fall within the $5/3$ to $2$ range, which suggests the presence of infinite variance in many of these $q$-Gaussian distributions, as the second moment diverges for $q>5/3$. Intriguingly, we observe a linear decreasing trend in $\beta$ versus distance to the sea, regardless of the detrending method used. This observation suggests that DO dynamics are influenced by spatial location, significantly impacted by geographical conditions, particularly proximity to the sea and associated seawater. There is greater variability or dispersion (larger $\beta$) in DO values for sites further away from the sea. This pattern may be due to the diminishing influence of seawater, which tends to stabilize oxygen levels closer to the coast. Factors such as decreased water mixing, reduced impact of tidal forces, and changes in salinity fronts, stratification, and sediment oxygen demand may also have a more pronounced influence inland.

Generally, the $\beta$ values from multiplicative detrending are bigger in magnitude than those deduced from additive decompositions. This suggests that the multiplicative method may be more sensitive to capturing complex or nonlinear variability in DO levels. This observation is consistent with our previously described findings in the section on Detrending. To understand the behavior and characteristics of extreme DO events more effectively, we formulate regression models in the next section.

\section*{Regression Analysis}
Understanding the statistical properties of extreme DO events, as explored in our previous section, is pivotal for developing robust predictive models. Using explainable machine learning techniques, we investigate and quantify the relationships between DO and other measurements of the water quality. Central to this effort is the identification of variables that significantly impact DO levels, which is essential for enhancing water quality management. Here, we have applied a gradient-boosted tree, specifically Light Gradient Boosted Machines (LightGBM or LGBM) \cite{ke2017lightgbm}, to the time-series data from TPut and TKB as examples, generating same-time predictions of the DO levels given a series of environmental and time features (see the Methods section for more details). SHAP values~\cite{lundberg2017unified,molnar2020interpretable}, approximating Shapley values, are then used to infer which features have the greatest effect on the generated regression models for each site.

All of the error results are displayed below in the form of symmetric mean absolute percentage error (SMAPE), computed as follows:
\begin{equation}
\text{SMAPE} (y_i, \hat{y}_i) = \frac{1}{n} \sum_{i=1}^{n} \frac{|y_{i} - \hat{y}_{i}|}{|y_{i}|+|\hat{y}_{i}|},
\end{equation}
where $n$ is the number of samples in the test set. $y_i, \hat{y}_i$ represent the real-time DO measurements and the LGBM predictions, respectively, for each corresponding entry $i$ within the test dataset.
As can be seen in Table.~\ref{tab:lgbm_mec}, the LGBM's performance, as measured by SMAPE, generally exhibits low error in predictions. Site TBGP displays the highest SMAPE, which may be attributable to its proximity to the sea, leading to more fluctuating and complex DO levels in water. TH, TChp and TBB exhibit relatively higher SMAPE values, likely due to the presence of the Mogden Sewage Treatment Works upsteam of the sites, resulting in more complex, changing pattern of DO levels heavily influenced by human activities.

\begin{table}[h]
    \centering
    \begin{tabular}{ |p{2cm}||p{2cm}|p{2cm}|p{3cm}|  }
        \toprule
        Site & LGBM & XGBoost & Linear Regression \\
        \midrule
        TPut & \textbf{2.8972} & 4.1564 & 5.8361 \\
        TPur & \textbf{2.5053} & 3.2652 & 3.5194 \\
        TKB & \textbf{3.8519} & 4.1534 & 4.5528 \\
        TH & 4.6179 & 6.0227 & \textbf{4.5480} \\
        TEB & 3.0311 & 3.2396 & \textbf{1.4791} \\
        TChP & \textbf{4.7982} & 5.6858 & 5.5984 \\
        TCaP & \textbf{3.6000} & 4.2287 & 5.3751 \\
        TBGP & \textbf{7.3040} & 5.6147 & 8.3671 \\
        TBB & \textbf{4.1096} & 4.1963 & 4.6866 \\
        \bottomrule
    \end{tabular}
    \caption{SMAPE values of the LGBM models compared to the values of the XGBoost and linear regression baseline models, for the nine Thames river sites. Results highlighted in bold indicate the best performance, with LGBM performing best in most cases.}
    \label{tab:lgbm_mec}
\end{table}

\begin{figure}[h]
  \centering
  \includegraphics[width=1\textwidth]{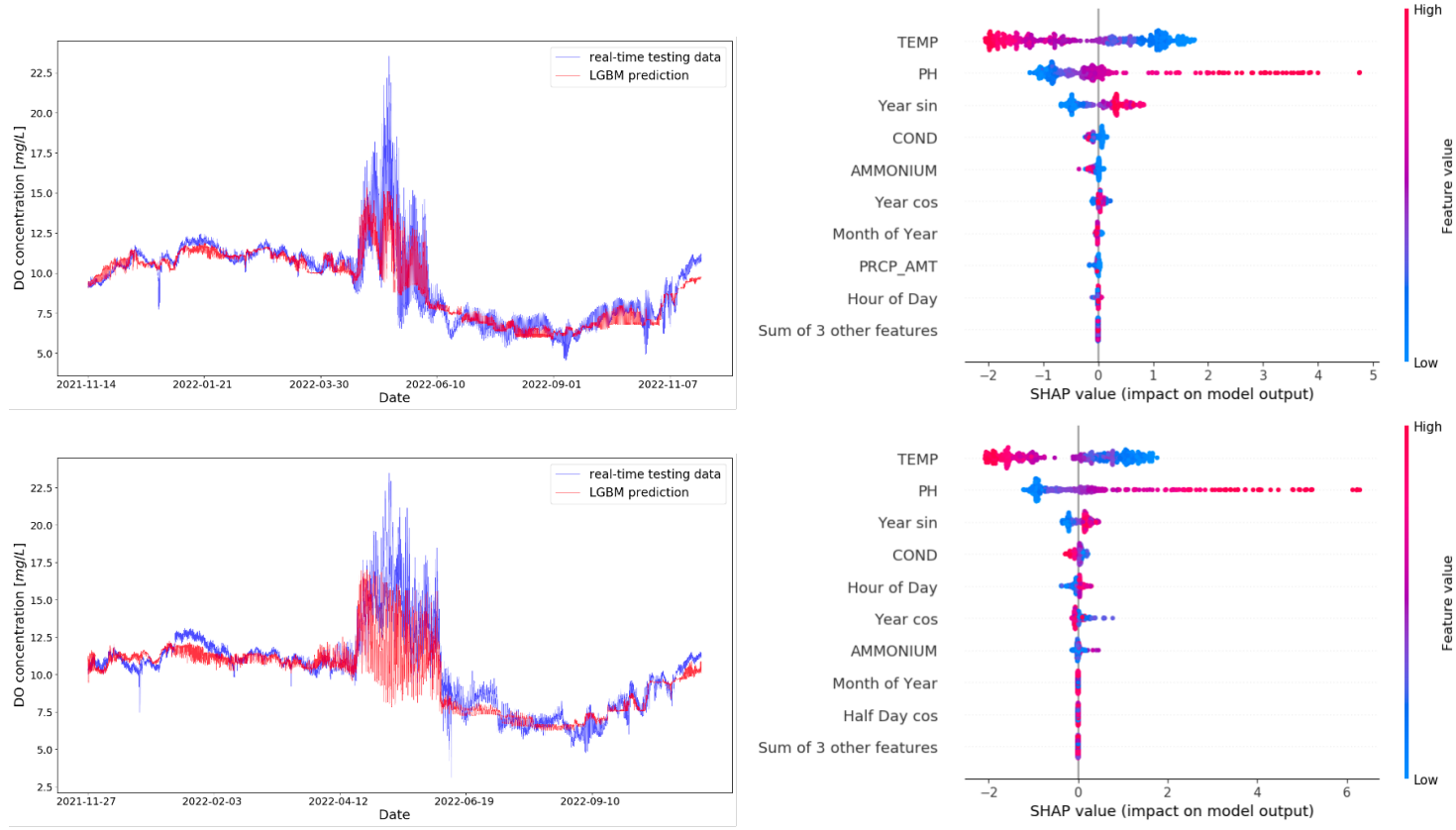}
  \caption{TPut (top) and TKB (bottom) data plotted against the same-time predictions by LGBM, all in chronological order. The SHAP values are plotted on the right hand side for each site, showing the calculated feature importances.}
  \label{fig:lgbm_comp}
\end{figure}

Although there is a lot of variability in the very short-time scale fluctuations in DO, Figure.~\ref{fig:lgbm_comp} shows that the generated regression models follow the peaks and troughs of DO variation over longer time spans, demonstrating that these models are utilizing the full range of possible and realistic DO values and are not sticking only to mean values. The excellent performance is attributable to several factors. LGBM utilizes leaf-wise (best-first) tree growth, which can lead to deeper and potentially more complex trees compared to the level-wise approach used by XGBoost. LGBM also discretizes continuous feature values into bins to speed up training and reduce memory usage. This means that LGBM is generally faster and more memory-efficient, particularly for large datasets, due to its histogram-based approach and optimizations like GOSS and EFB.

We trained the model to get an understanding of how the different water quality indicators interact or work together. Fig.~\ref{fig:lgbm_comp} demonstrates the calculated feature importances of the featured models. Negative SHAP values indicate that a feature leads to lower DO predictions, whereas positive SHAP values indicate that a feature pushes the prediction towards higher DO levels. The colour represents the feature value, ranging from high (red) to low (blue). This information effectively increases our understanding of how the different measurements interact and work together: We can see that the order and correlation of feature importance is very similar for both sites, with the top three features being Temperature, pH and sine of the Year (encoding the time of the year). The other sites also show similar results with 4 out of the 7 other sites having the same order for the top three features. This suggests that for future research, temperature and pH sensors should be prioritised whether for analysis or predicting DO. Temperature is the most important in 6 out of those 7 sites and ranked second in the other one. Temperature and DO have a negative correlation, i.e. higher values of temperature reduce DO, while lower values increase DO. Meanwhile, pH, the second most impactful factor, and the predicted DO values exhibit a positive correlation. Specifically, alkaline waters are associated with an increase in the predicted DO values. Conversely, acidity tends to decrease the DO prediction. This is consistent with the fact that high-pH water contains low levels of hydrogen ions, meaning more oxygen is available, and vice versa. From the selected calendar features, the Year (and its encoding via sine) clearly takes precedence, suggesting that the most predictable fluctuations are over an annual cycle of the DO concentration.

The feature ranking deduced here for DO differs from a previous feature importance analysis of electrical conductivity in the river Chess~\cite{schafer2022machine}, where temperature was not as important, and River Level, pH, and Month were the most significant features. This can be explained by the fact that DO is highly dependent on water temperature, and these annual fluctuations are the primary control on DO levels. Furthermore, pH varies with the photosynthetic activity of algae, providing a biological control on water quality. During daylight, photosynthesis increases oxygen production, reduces carbon dioxide (and thus carbonic acid) levels, and raises alkalinity. Conversely, at night, increased respiration raises carbon dioxide levels, lowers DO, and increases water acidity. Additionally, monthly and yearly periodicities were prioritized when predicting electrical conductivity and DO, respectively. This distinction highlights that although both rivers are subject to temporal factors, the specific impacts of these factors—monthly and yearly periodicities—on water quality indicators are distinct. The 'month' feature indicates that electrical conductivity is more affected by monthly patterns due to factors such as heavy rainfall or dry spells. Conversely, the 'time of the year' feature implies that DO levels are more influenced by annual patterns, likely due to temperature variations and related photosynthetic activities of aquatic plants. Another possible reason for the differences is the greater influence of anthropogenic activities, such as sewage pollution, in the River Thames compared to the more remotely located (small) River Chess. 

\section*{Time series forecasting}
In the previous section, we focused on the analysis and interaction of different quantities (e.g. DO and temperature) at the same time. Now, we are moving to the question of whether and how an important water quality indicator, such as DO, can be forecasted for the next hours or even days. In this section, we briefly review and apply the recently developed Informer model~\cite{zhou2021informer}, which is a state-of-the-art deep learning model that can be employed to predict a wide range of long-sequence time series, including DO. To our knowledge, this marks (one of) the first application(s) of this model to sequential DO forecasting, aimed at achieving precise forecasts for multiple time steps.

The core of the Informer model is a stacked transformer architecture, which includes an encoder and a decoder. The encoder captures dependencies in the input data, transforming the input sequence into a set of high-level features. Subsequently, the decoder takes the encoded sequence and generates the output sequence. In addition to the Informer, we compare various forecasting models, including baseline approaches such as "Last", "Repeat", and "Linear", as well as neural network architectures like Sequential Dense Network ("Dense")~\cite{huang2017densely}, Convolutional Neural Network ("Conv")~\cite{o2015introduction}, and Long Short-Term Memory ("LSTM")~\cite{hochreiter1997long}. Details regarding the architecture and hyperparameter settings of these models are provided in the Methods section. 

To evaluate the performance of a model, we compute its mean absolute error (MAE) and SMAPE for forecasts $ t \in {1, 12, 24, 48} $ quarter-hours into the future based on the test sets. They are computed as follows:
\begin{equation}
MAE_t (y_i, \hat{y}_i) = \frac{1}{n} \sum_{i=1}^{n} |y_{i,t} - \hat{y}_{i,t}|,
\end{equation}
\begin{equation}
SMAPE_t (y_i, \hat{y}_i) = \frac{1}{n} \sum_{i=1}^{n} \frac{|y_{i,t} - \hat{y}_{i,t}|}{|y_{i,t}|+|\hat{y}_{i,t}|},
\end{equation}
where $n$ is the number of samples in the test set. $y_i, \hat{y}_i$ represent the actual DO measurements and the DO predictions respectively, for each prediction horizon $t$ and corresponding entry $i$ within the test dataset. The metrics were computed for five iterations and then averaged to evaluate each model’s performance in prediction compatibility. The analysis was performed on the TBGP dataset, and comprises fourteen covariates. The results are presented in Table~\ref{fig:inf_performance}.

\begin{table}[h]
\centering
\begin{tabular}{|l|l|r|r|r|r|r|r|r|}
\toprule
Input\_Pred &   Metrics &    Last &  Repeat &  Linear &   Dense &    Conv &    LSTM &                   Informer \\
\midrule
      48\_1 &       MAE &  \textbf{0.0758} &  0.2524 &  0.0833 &  0.1078 &  0.1255 &  0.0801 &  0.1143 \\
           & SMAPE [\%] & \textbf{12.2542} & 31.6322 & 14.7789 & 19.2055 & 21.3571 & 13.4692 & 16.3717 \\\hline      
     48\_12 &       MAE &  0.1900 &  0.1780 &  0.1884 &  0.1997 &  0.2144 &  0.1828 &        \textbf{0.1596} \\
           & SMAPE [\%] & 25.2479 & 24.0496 & 24.5739 & 26.3096 & 27.8113 & 24.8176 &          \textbf{21.7630} \\\hline
     48\_24 &       MAE &  0.2402 &  0.1779 &  0.2670 &  0.3134 &  0.3293 &  0.1953 &            \textbf{0.1755} \\
           & SMAPE [\%] & 30.3745 & 24.0608 & 31.8956 & 36.3300 & 37.6673 & 25.4730 &  \textbf{23.5131} \\\hline
     48\_48 &       MAE &  0.2528 &  0.1777 &  0.2911 &  0.3474 &  0.3394 &  0.1788 & \textbf{0.1753} \\
           & SMAPE [\%] & 31.7033 & 24.0704 & 34.2229 & 39.0919 & 38.6273 & 25.4387 &  \textbf{23.5828} \\\hline
      96\_1 &       MAE &  \textbf{0.0758}&  0.2506 &  0.0760 &  0.0854 &  0.0885 &  0.0800 & 0.1076 \\
           & SMAPE [\%] & \textbf{12.2542} & 31.4513 & 12.3700 & 15.1089 & 16.0967 & 13.7094 &  15.5393 \\\hline
     96\_12 &       MAE &  0.1902 &  0.1781 &  0.1886 &  0.1992 &  0.2112 &  0.1799 &      \textbf{0.1457} \\
           & SMAPE [\%] & 25.2728 & 24.0638 & 24.6089 & 26.3057 & 27.5138 & 23.6358 &        \textbf{21.0986} \\\hline
     96\_24 &       MAE &  0.2404 &  0.1780 &  0.2681 &  0.2959 &  0.3128 &  0.2048 &        \textbf{0.1531} \\
           & SMAPE [\%] & 30.3912 & 24.0725 & 36.8220 & 35.5253 & 38.0492 & 33.0254 &         \textbf{22.2273} \\\hline
     96\_48 &       MAE &  0.2531 &  0.1778 &  0.2919 &  0.3152 &  0.3190 &  0.1739 &        \textbf{0.1569} \\
           & SMAPE [\%] & 31.7269 & 24.0938 & 34.3356 & 36.8165 & 37.2676 & 24.9971 &         \textbf{22.2918} \\\hline
     192\_1 &       MAE &  \textbf{0.0758} &  0.2597 &  0.0760 &  0.0855 &  0.0859 &  0.0815 & 0.1060 \\
           & SMAPE [\%] & \textbf{12.2542} & 32.5443 & 12.3848 & 15.2599 & 14.7940 & 13.8408 &  15.9681 \\\hline
    192\_12 &       MAE &  0.1908 &  0.1790 &  0.1893 &  0.1996 &  0.2198 &  0.1825 & \textbf{0.1348} \\
           & SMAPE [\%] & 25.3368 & 24.1895 & 24.7196 & 26.3322 & 28.6294 & 24.0999 &  \textbf{20.0625} \\\hline
    192\_24 &       MAE &  0.2412 &  0.1789 &  0.2685 &  0.2984 &  0.3127 &  0.2121 &       \textbf{0.1430} \\
           & SMAPE [\%] & 30.4682 & 24.1993 & 32.1065 & 35.6741 & 36.8389 & 27.9968 &  \textbf{21.3131} \\\hline
    192\_48 &       MAE &  0.2540 &  0.1788 &  0.2923 &  0.3157 &  0.3193 &  0.1750 &       \textbf{0.1495} \\
           & SMAPE [\%] & 31.8114 & 24.2165 & 34.4026 & 36.8958 & 37.4802 & 24.4682 &  \textbf{21.9554} \\\hline
\end{tabular}
\caption{MAE and SMAPE results for sequence time-series forecasting across various horizons ("Pred") and input lengths ("Input") on the TBGP data set using different methods. The best results are highlighted in bold.}
\label{fig:inf_performance}
\end{table}

\begin{figure}[ht!]
    \centering
    \includegraphics[width=1\textwidth]{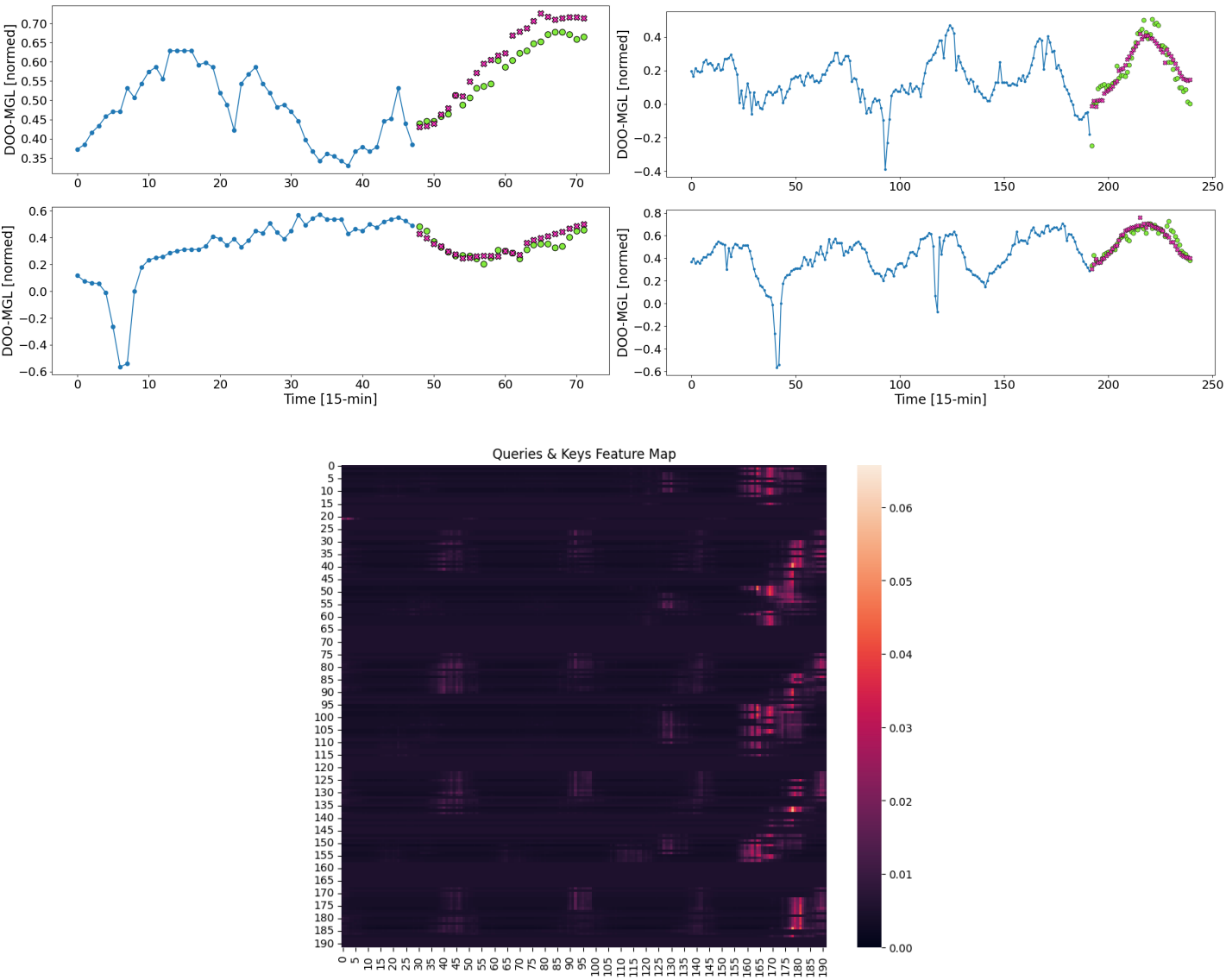}
    \caption{Top: Example windows for the Informer's forecast of $t\in{12, 48}$ with input lengths of $48$ and $192$ time steps, respectively. The blue line represents the DO input at every time step. While the model processes all features, this visualization only displays the DO. The green dots depict the desired prediction values. The purple crosses represent the forecast of DO made by the Informer model in a single shot. Bottom: A visualization of the attention weights, represented through a heatmap, when predicting 48 future time steps from 192 past steps. The $x$-axis and $y$-axis correspond to the keys and queries, respectively, associated with each element in the sequence. Lighter colors correspond to higher weights.}
    \label{fig:192_48_fm}
\end{figure}

By comparing the seven models, the MAE and SMAPE averaged over all times and outputs from the Informer are the lowest, except when predicting for a single time step. This demonstrates the superior predictive capability of the Informer, particularly for long prediction horizons, compared to the other models. Notably, the performance of the Informer in predicting $12$, $24$ and $48$ time steps (3, 8 and 12 hours respectively) improves with an increased input sequence, as indicated by the decreasing MAE and SMAPE. This can be attributed to the novel Probsparse self-attention mechanism of the Informer model, which dynamically focuses on the most informative parts of a long input sequence and effectively captures long-range dependencies and temporal dynamics. The Repeat model appears to be the second-best for forecast horizons of $t \in {12, 24, 48}$. This can be attributed to the half-day periodicity of DO, which the Repeat model utilizes, aligning well with short-term dynamics in the aquatic environment and thereby enhancing its forecasting ability. Example windows for Informer's forecasting results of $t\in{12, 48}$ are illustrated in Fig.~\ref{fig:192_48_fm} top. The Informer model exhibits capabilities in forecasting future values of DO for both shorter (24) and longer (48) time steps, with the predicted values generally aligning with the actual values.

Above, we demonstrated the high-performance capabilities of the Informer. However, which inputs are particularly important remains unknown so far. We will next explore the self-attention mechanism that underlies these models, aiming to identify the elements driving their forecasting power. The self-attention mechanism, which assigns different weights to different inputs for the prediction ("giving attention to the important contributors"), is detailed in the Methods section.

To identify which parts of the input are important for the prediction, we illustrate the attention weights in scenarios involving the maximum input and output lengths, i.e., 192 input time steps (48 hours) to predict 48 time steps (12 hours) in Fig.~\ref{fig:192_48_fm}. This approach offers insights into the internal workings of the model's self-attention mechanism. Note that the last few columns on the right contain numerous high attention weights. The information at positions $175$ to $180$, representing the recent time steps in the input sequence, was highly influential during two distinct periods: morning to early afternoon and late evening to early morning. Each of these periods spans 6 hours. This observation suggests that our model has effectively incorporated half-daily patterns while creating new representations. Furthermore, the high attention weights in columns $160$ to $170$ indicate that the model paid significant attention to the keys from the 16th to the 26th time step of the last half-day while processing the queries. This aligns with the Repeat model's good performance in forecasting results. The identified high-risk periods, specifically from morning to early afternoon and from late evening to early morning of the most recent time steps in the input, as well as the 16th to 26th time steps of the last half-day, are crucial. When combined with accurate long-term forecasting, these insights could guide and refine environmental governance strategies. Such proactive measures in water management may include controlled treatment releases and oxygen injection, particularly when forecasting low DO levels.

\section*{Discussion}
Our research extends previous results, which were obtained for a comparatively small river, the River Chess~\cite{SCHAFER2021102881}, to a much larger river, the tidal stretch of the River Thames, covering nine sites over a distance of 45 km. The study for the River Chess~\cite{SCHAFER2021102881} utilized the additive EMD method for detrending, with fluctuations following a $q$-Gaussian distribution. In contrast, the time series from the River Thames exhibit more pronounced tidal periodic variations, varying significantly in frequency and amplitude. To address this difference, we employed the multiplicative detrending method for the River Thames, which proved to be more effective, as demonstrated by the higher log-likelihood values obtained from the $q$-Gaussian fittings compared to those from the River Chess. After detrending, the fluctuation statistics, in very good approximation, are $q$-Gaussian and thus exhibit power-law tails. This pattern is confirmed across all sites along the tidal Thames. Moreover, our analysis did not only focus on individual sites but also systematically compared the best-fitting superstatistical parameters across all sites. This comparative analysis helped to derive correlations between these parameters and their distance to the sea, highlighting how geographical conditions, particularly proximity to the sea and associated seawater influence, affect the tail behaviors of the PDFs of DO. 

In the context of machine learning, previous research for the River Chess \cite{schafer2022machine} demonstrated the effectiveness of the LGBM in predicting temperature and electrical conductivity. We expanded the use of LGBM to include dissolved oxygen in the River Thames, emphasizing the robustness of the boosted tree approach across different water quality indicators and rivers. Our feature importance analysis revealed differences in the importance of and (inter)dependencies between individual indicators such as electrical conductivity as compared to the River Chess~\cite{schafer2022machine}, where temperature was less significant compared to river level, pH, and month. Although the Transformer model has been recently applied to forecast the river WQI in the Bhavani River, which combines various water quality parameters to provide an overview of water quality \cite{nair2023temporal}, we aimed to forecast individual indicators such as DO. This approach offers detailed insights into specific aspects of water quality, enabling targeted investigations. In comparison, we demonstrated that the Informer model is an effective tool for accurate long-term forecasts of DO up to a half-day, achieving the lowest SMAPE and MAE. The Informer's embedded ProbSparse self-attention mechanism highlighted specific time periods—morning to early afternoon, late evening to early morning in the most recent inputs, and the 16th to 26th quarter-hour of the previous half-day—as particularly influential in the forecasting process, offering a significant and novel insight.

Our data analysis so far is limited by the relatively small number of real-time measuring sites along the River Thames. Further data sets from more sites located at each river stretch could help to provide a more systematic picture of the overall statistics, allowing us to judge which PDF shapes are generic and which observed shapes are specific to a given measuring site. Currently, the resolution discrepancy between our rainfall data and other environmental parameters has limited its utility in predictive models. High-resolution rainfall data in future studies could refine our analysis, potentially increasing the accuracy of our forecasts. 

Our research may provide policymakers and other stakeholders with insights and effective forecasting tools to effectively monitor and manage the health of aquatic ecosystems. The Informer model could be integrated into a monitoring system to alert environmental agencies when forecasted values indicate potential risks, such as low oxygen levels which could harm aquatic life. Additionally, our analysis identified critical periods—specifically from morning to early afternoon, and from late evening to early morning—that, along with key factors like temperature, pH, and time of year, can inform and update environmental regulations and policies. For instance, implementing measures such as oxygen injection or controlled water treatment releases during these high-risk periods could effectively enhance environmental protection efforts for the River Thames. Furthermore, correlation analysis between the superstatistical parameters and distance to the sea reveals that inland sites, which are more susceptible to greater variability or dispersion in DO values, could be prioritized for enhanced monitoring. These sites could be equipped with more precise sensor technology to better reflect changes in river conditions. In this context, the method developed in our study provides a valuable tool for researchers and policymakers in modeling the tail behaviors (extreme events) of environmental indicators.

Our methods developed here could pave the way for broader exploration and application. This possible extension may include different environments ranging from coastal to estuarine, and of course also rivers in other countries than the UK. Future studies may extend this methodology also to other observables, such as electrical conductivity, building on the insights gained from our targeted examination of DO, and examining cross-correlations between different water quality indicators. While our focus has been on the statistical properties of time series data, future work will also consider the specific geographical and environmental conditions of each river site. This approach will enable more targeted actions and scientific investigations to enable, maintain and preserve the health of river ecosystems. For comparative analysis of machine learning models, additional statistical methods could be explored to aid in choosing the best-performing model. In addition to comparing performance metrics in DO estimation, recent research in the Sakarya Basin, Turkey~\cite{10.1007/978-3-031-43169-2_21} utilized Taylor and violin diagrams to provide a more comprehensive statistical comparison of model performances. The use of Taylor diagrams, error boxplots, and violin graphs can be an additional tool to further enhance the visualization of model performance comparisons. Moreover, the Kruskal-Wallis test could provide a statistical basis for evaluating and validating the performance of the different models in future studies. Forecasting methods such as Temporal Fusion Transformer~\cite{lim2021temporal} and N-HiTS~\cite{Challu_Olivares_Oreshkin_Garza} have the potential to enhance predictions. Probabilistic Transformer~\cite{NEURIPS2021_c68bd905} and Natural Gradient Boosting~\cite{pmlr-v119-duan20a}, on the other hand, allow for forecasts and probabilistic predictions respectively, with uncertainty estimates. However, these methods require higher computational cost and a different experimental design, which were beyond the resources and scope of this paper. We prioritized a first proof-of-concept application and evaluation of these existing models to establish an accurate DO prediction technique. Future projects could explore these methodologies to further investigate the impact of uncertainty on ML model predictions.

\section*{Conclusion}
This study introduces a novel approach to analyzing and forecasting the spatio-temporal dynamics of DO in the River Thames by applying superstatistical methods and machine learning models. We introduced a multiplicative detrending method that effectively separates trends from rapidly fluctuating DO levels, leading to more accurate $q$-Gaussian fittings. The fluctuations were observed to follow a $q$-Gaussian distribution and exhibit power-law tails, confirming previous results obtained for much smaller rivers. The superstatistical parameters $\beta$ and $q$ of the $q$-Gaussians revealed that proximity to the sea and associated seawater significantly influences DO fluctuations. Another key contribution is demonstrating the robustness of the LGBM model for real-time DO predictions and the superior long-term forecasting capabilities of the Informer model. Interpretation of the LGBM model using SHAP values consistently identified temperature, pH, and time of the year as dominant predictors of the DO fluctuations across different sites. Additionally, an exploration of the Informer model’s self-attention mechanism revealed that inputs from morning to early afternoon and from late evening to early morning of the most recent time steps, as well as from the 16th to the 26th quarter-hour of the previous half-day, primarily drive its superior performance.

Our study is limited by the small number of real-time measuring sites and the resolution discrepancy between rainfall data and other environmental parameters. Future research should address these limitations by incorporating more extensive data. Furthermore, future studies may extend our methodology to other observables, such as electrical conductivity, in different river systems outside the UK, and conduct closer examinations of geographical and site-specific extreme events at each location. While our current paper provides a proof-of-concept approach to DO forecasting, future studies should explore advanced models like the Temporal Fusion Transformer and Natural Gradient Boosting to enhance accuracy and account for uncertainty. Our findings suggest that the Informer model could be integrated into monitoring systems to provide early warnings for low DO levels, identifying critical time windows when targeted environmental interventions could be implemented. The $q$-Gaussian method for analysing extreme events and tail behaviour of PDFs can also be adapted for different contexts, offering valuable quantitative tools for identifying sites that exhibit more extreme variations than others and that may thus require enhanced monitoring.

\section*{Methods}
\subsection*{The data set considered}
We utilise water quality data from nine monitoring sites along the River Thames from Meteor Communications telemetry system~\cite{Meteor}. The Environmental Agency (EA), in collaboration with Meteor Communications, operates a sensor web network across the River Thames. This network features multi-parameter water quality sondes connected to Meteor's telemetry system, transmitting nine measurement data in real-time to a central database, accessible via a web host. Water quality indicators were measured using real-time sensors from 01/12/2017 to 01/12/2022 with a temporal resolution of 15-min. We use the following water quality indicators:
\begin{enumerate}
\item Dissolved oxygen (DO) refers to the concentration of oxygen gas incorporated in water, vital for aquatic life, measured in milligrams per liter (mg/L).
\item Temperature (Temp) is a local temperature measured in degrees Celsius (°C).
\item Electrical conductivity denotes the ability of a material to conduct an electrical charge, measured in microsiemens per centimeter (µS/cm).
\item pH (pH) indicates the acidity or alkalinity of a solution, using the pH scale ranging from 0 to 14.
\item Ammonium (AMMONIUM) refers to the concentration of ammonium ions, a measure of nitrogen pollution in water, measured in milligrams per liter (mg/L).
\item Turbidity (TURBIDITY) refers to the cloudiness or haziness of a fluid caused by large numbers of individual particles, measured in nephelometric turbidity units (NTU).

Additionally, we extract rainfall data from the CEDA Archive~\cite{midas_daily,midas_hourly} to take into account the impact of rainfall on water levels in our predictions and forecasts. We pair each site's data set with daily/hourly rainfall data from the closest current running rainfall monitoring station. For example, we extract daily rainfall data from the site "DEPTFORD P STA", which is the closest current running rainfall monitoring station to the site TBGP (4.5 km aerial distance). We paired the daily rainfall data with the 15-minute measurements taken on the same day to incorporate its potential impact as a predictive feature in the forecast. 
\end{enumerate}

\subsection*{Data processing}
The data sets retrieved contain several problems, which we rectify in the following way:
\begin{enumerate}
    \item  DO and DO-MGL parameters not considered, instead use the DOO-MGL parameter measured with an optical optode for more reliable data. 'DOO-MGL' is equivalent to the commonly known 'DO' in this paper.
    \item  DOO-MGL measurements exceeding $25 mg/L$ are deemed faulty and discarded, as according to Wetzel~\cite{wetzel2001limnology}, dissolved oxygen in most natural waters typically ranges between $8-14 mg/L$ under standard pressure. While this can increase based on temperature, salinity, and pressure interactions, achieving a concentration as high as $25 mg/L$ is generally unlikely under standard conditions.
    \item Remove 'salinity' measurements as they are the same as electrical conductivity.
    \item The sites experienced unusual fluctuations in DOO-MGL in August 2022 due to the artificial injection of oxygen into the River Thames~\cite{Murugesu2023Thames}. To eliminate bias from human interaction, we have excluded these measurements from our analysis.
    \item We remove non-positive measurements for 'COND'(electrical conductivity), 'PH', 'AMMONIUM', 'Turbidity' and 'DOO-MGL' indicators.
    \item Remove large spikes and sudden dropouts due to probe faults or human activities, such as the oxygen pumped into the river from a boat in August 2022.
    \item EC measurements in ms/cm unit, change to us/cm.
    \item Remove data outages, which were caused by sonde changes.
   
\end{enumerate}

We measure the aerial distances between the sites and the Thames estuary in kilometers (dist\_to\_sea), to analyse the effect of the sea distance on the results.The estuary of the Thames is the transitional area where the River Thames's fresh waters mix with the saltwater of the North Sea. 

\begin{figure}[H]
  \centering
  \includegraphics[width=8.5cm]{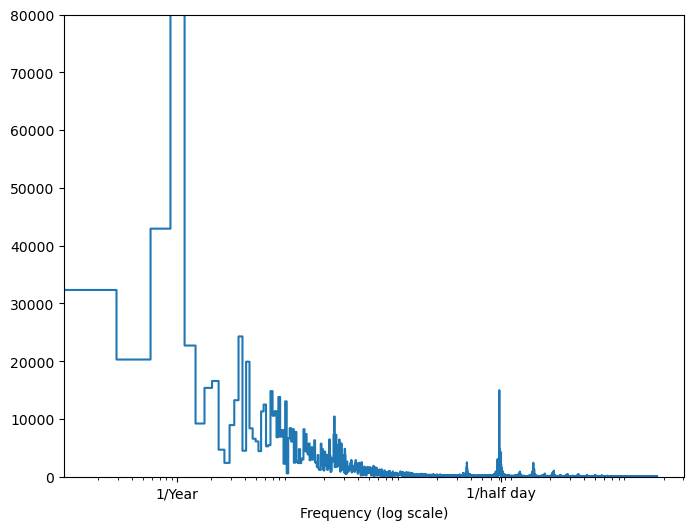}
  \caption{By performing a Fast Fourier Transform, we can determine the important frequencies. Note the intuitive peaks at frequencies near 1/year and 1/half day.}
  \label{fig:FFT}
\end{figure}

Other covariates include time and calendar features: the hour of the day, the day of the week, the month of the year (all encoded as value between [-0.5, 0.5]), the time of the half-day and the time of the year (both sine- and cosine-encoded). The latter two signals are extracted specifically to manage both half-daily and yearly periodicity. We identify these as the two most critical frequencies, accomplished by extracting features using the Fast Fourier Transform, see Fig.~\ref{fig:FFT}. In total, this amounts to fourteen covariates. All the models employ a multivariate feature set to predict a single output feature, DO, across all output time steps. 

We use a (80\%, 20\%) split for the training and test sets for the regression analysis, and a (70\%, 20\%, 10\%) split for the training, validation, and test sets for time series forecasting. Note that the data is not being randomly shuffled before splitting for two reasons. Firstly, it allows us to ensure that chopping the data into windows of consecutive samples is still possible. Secondly, it ensures that the results obtained from validation and testing are more realistic as they are being evaluated on data that was collected after the model was trained. Normalising the data is the next step, and the mean and standard deviation are calculated based on the training set. Thus, the models are not exposed to data from the validation and test sets.

\subsection*{Detrending methods}
We implement seasonal decomposition using the python statsmodels.tsa.seasonal package~\cite{statsmodels2023}, and EMD using the PyEMD package~\cite{pyemd}. We choose a filtering frequency of $f=6$ hours and vary the number of modes dropped based on each site's trajectories, as illustrated in the code.

\subsection*{Regression models}
\begin{figure}[H]
  \centering
  \includegraphics[width=10cm]{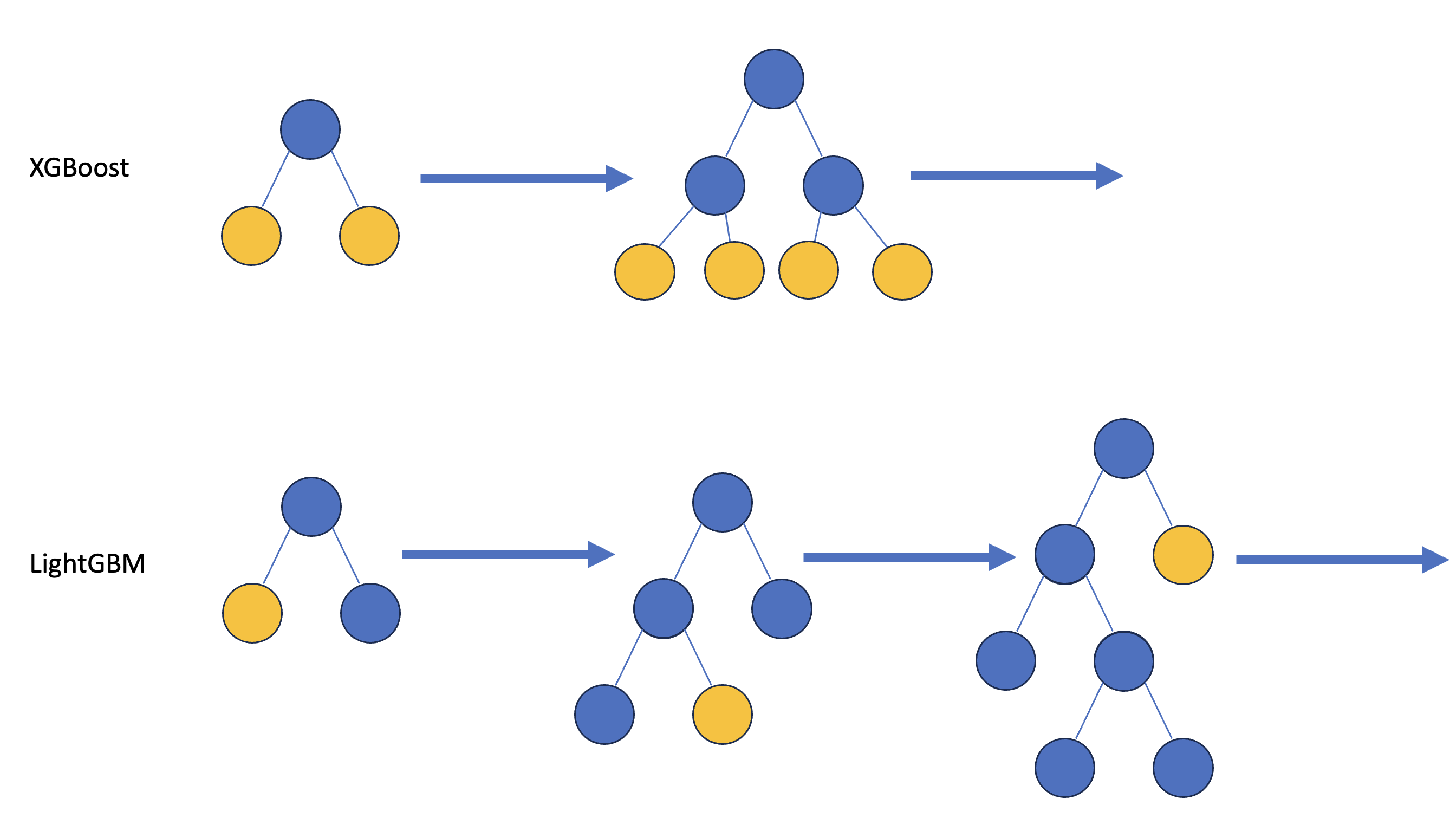}
  \caption{Illustration of the LGBM model. Decision tree is built leaf by leaf as opposed to other methods such as XGBoost gradient-boosted trees which build level by level.}
  \label{fig:lgbm}
\end{figure}

A Light Gradient Boosting Machine (LGBM)~\cite{ke2017lightgbm} was used as the interpretable regression model for the river data. This is a gradient-boosting method which grows gradient-boosted trees leaf by leaf rather than row by row and uses Gradient-Based One-Sided Sampling (GOSS) and Exclusive Feature Bundling (EFB) to allow its histogram-based decision tree learning algorithm to run faster, see Fig.~\ref{fig:lgbm}.

To tune the hyperparameters of the LGBM model, the Optuna tuning software \cite{akiba2019optuna} was used (through the Verstack wrapper), a stepwise algorithm to tune hyperparameters.

To interpret the results of the LGBM models, a random sample of 500 SHAP (Shapley Additive Explanations) was used~\cite{lundberg2017unified,molnar2020interpretable}. This is a method from coalitional game theory that tells us how to fairly distribute the "payout" (the prediction by the model) between the "players" (the features in the model) in a "coalition" (the model). The explanation itself is 
\(g(z')=\phi_0+\sum_{j=1}^M\phi_jz_j' \)
where g is the explanation model, 
\(z'\in\{0,1\}^M \)
is the coalition vector, M is the maximum coalition size and 
\(\phi_j\in\mathbb{R} \)
is the feature attribution for a feature $j$, the Shapley values.

\subsection*{Forecasting models}
\begin{figure}[H]
  \centering
  \includegraphics[width=10cm]{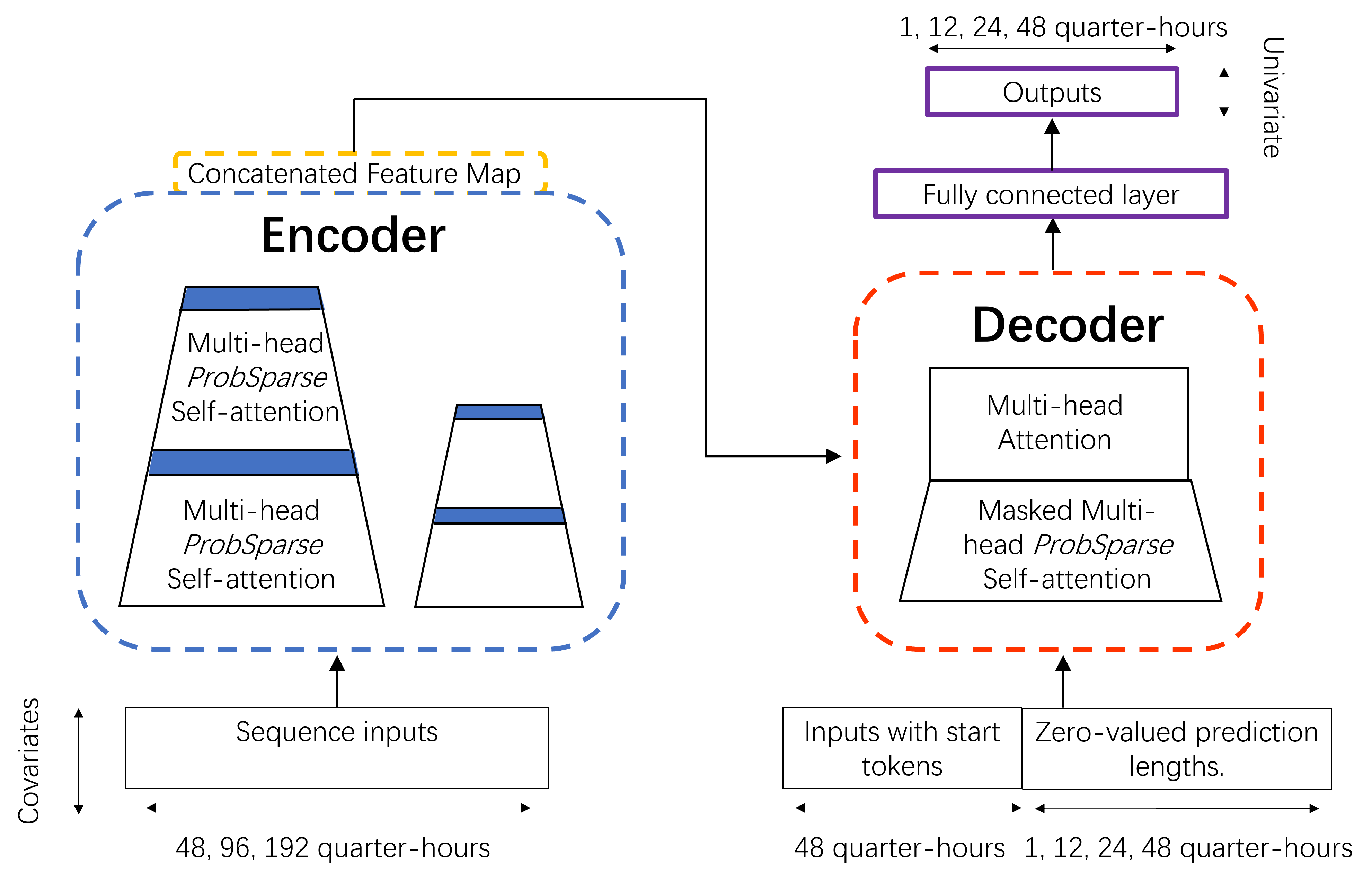}
  \caption{Informer model architecture.}
  \label{fig:inf_mec}
\end{figure}

The employed forecasting models are as follows:
\begin{enumerate}
    \item{Last: A simple baseline is to repeat the last input time step for the required number of output time steps.}
    \item{Repeat: The second baseline method involves repeating the same length of initial data from the previous half-day as the prediction length. This assumes that the patterns from the start of the previous half-day will be similar in the upcoming period, given the half-day periodicity observed in our data.}
    \item{Linear: A linear model is based on the last axis of the data, changing its format from [batch, time, inputs] to [batch, time, units]. It works independently on each item across both the batch and time axes. The model aims to predict time steps based on a single input time step using a linear projection.}
    \item{Sequential Dense Network (Dense): The model selects the last time step from the input, reshaping it from [batch, time, features] to [batch, 1, features]. It then passes the data through a dense layer of 512 units with 'ReLU'(Rectified Linear Unit) activation~\cite{agarap2018deep}, maintaining the shape as [batch, 1, dense units]. Finally, it reshapes the output to match the desired number of steps and features. It is capable of capturing more complex patterns in the data than a linear model.}
    \item{Convolutional Neural Network (Conv)~\cite{726791}: The model first slices the last three time steps from the input, changing the shape from [batch, time, features] to [batch, 3, features]. It then applies a 1D convolutional layer with 256 units and 'relu' activation, reshaping the data to [batch, 1, conv units]. A dense layer and a reshape layer adjusts the data to [batch, out steps, 1], yielding a convolutional model capable of producing a prediction based on that sequence. }
    \item{Long Short-Term Memory (LSTM)~\cite{hochreiter1997long}: LSTM network, a recurrent neural network, processes a time series in a sequential manner and maintains an internal state across time steps. It starts with an LSTM layer containing 32 units, transforming the data shape from [batch, time, features] to [batch, lstm units]. After experimentation with various numbers of layers and units per layer, we found this model architecture provided the best performance. If historical inputs are relevant for the predictions, the LSTM model can be trained to exploit them effectively. In this specific application, the model accumulates internal state information for a half-day before generating a forecast for the subsequent time steps.}
\end{enumerate}
Additionally, we use the Informer, which is the cutting-edge machine learning methodology for numerous sequence-oriented tasks. Zhou et al.\cite{zhou2021informer} introduced a unique "ProbSparse Self-Attention" mechanism that targets the main influential factors while maintaining a sparse attention distribution. This mechanism significantly reduces computational complexity by probabilistically selecting the most important elements from the input sequence to pay attention to. An overview of the mechanism behind informer can be found in the Fig.~\ref{fig:inf_mec}. Left: The encoder takes in 14-dimensional input sequences for 48, 92, and 192 time spans. ProbSparse self-attention selects the active queries. Trapezoids represent the self-attention distillation operation, which reduces the network size by extracting dominant attention. Right: The decoder acquires the token length of the preceding 48 time steps of input, fills the prediction elements with zeros, examines the structure of the attention feature map. In a single shot, it generates univariate output predictions for time spans of 1, 12, 24, and 48.

The data fed into the model is structured with the shape [batch, time, features]. Here, we generate batches of 32 windows from the training, evaluation, and test data. The order of the windows is shuffled before they are batched, which in turn means that the order of the data within each batch is also random. In each batch, the model makes a set of independent one-shot predictions on future time steps. The term 'time' represents the sequence length per sample. We use values from half a day (i.e., 48 time steps), a full day, or two days as input, with forecasting horizons of 1, 12, 24, and 48 quarter-hours. The term 'features' refers to the number of attributes each time step possesses. The output shape of the model is [batch, time, 1], which reflects the prediction of only one feature. All the models employ a multivariate feature set to predict a single output feature, DO, across all output time steps. 

We utilise vectors representing the previous 48 time steps (equivalent to half a day) as input to the encoder, and vectors representing the previous 24 time steps for the start token length of the decoder. The prediction sequence length are {1, 12, 24, or 48} time steps. The architecture includes three encoder and three decoder layers, with each attention module having eight heads. The model dimension is set to 512. The specific hyperparameters of the model are provided in the code. Each model undergoes training utilising the MAE as the loss function. Our proposed methods are optimised using the Adam optimiser~\cite{kingma2014adam}. The learning rate is set to 0.0001, and this is subsequently halved with each increment in the epoch number. We employ early stopping to attain the minimal generalization error on the validation data, with a patience of 3 epochs. In the evaluation, the model with the lowest validation error is saved.

In the Informer model, the encoder takes in the complete input sequence and generates a corresponding sequence of representations. Each input requires three distinct representations, known as the key (K), query (Q), and value (V). Each representation is a vector encoding information about a specific timestep in the input, considering the context of the entire sequence. The decoder receives these representations to generate the output sequence. To derive these representations, every input is multiplied with a distinct set of weights assigned for K, Q, and V. These weight matrices represent trainable parameters of the model, which are optimised during training to minimise the loss function. Subsequently, the dot product of the query vector and every key vector in the sequence is computed to generate attention scores. These scores determine the level of attention each element should receive. These results are subjected to a softmax function, converting the scores into probabilities, known as attention weights. The final output is a composite of these interactions and corresponding attention scores. The ProbSparse Self-attention can be computed as follows:
\begin{equation}
\text{Attention}(Q, K, V) = \text{softmax}(\overline{Q}K^T/\sqrt{d})V,
\end{equation}
where $d$ represents the dimensionality of the queries and keys, while $\overline{Q}$ is the sparse query matrix containing the top-u queries. This configuration allows each key to attend only to the 'u' most dominant queries. The aim is to select the most "active" queries, those that result in high attention weights. The softmax function ensures the weights sum to $1$ and are each between $0$ and $1$.

\subsection*{Data availability}
The code to generate the figures in the paper, as well as the implementation of the method for the data sets used in the paper, are available at \url{https://github.com/hurst0415/Water-quality-analysis-for-the-Thames-River}.

\bibliographystyle{naturemag}

\begin{thebibliography}{10}
\expandafter\ifx\csname url\endcsname\relax
  \def\url#1{\texttt{#1}}\fi
\expandafter\ifx\csname urlprefix\endcsname\relax\def\urlprefix{URL }\fi
\providecommand{\bibinfo}[2]{#2}
\providecommand{\eprint}[2][]{\url{#2}}

\bibitem{mccormick2021state}
\bibinfo{author}{McCormick, H.}, \bibinfo{author}{Cox, T.},
  \bibinfo{author}{Pecorelli, J.} \& \bibinfo{author}{Debney, A.}
\newblock \bibinfo{title}{The state of the thames 2021: Environmental trends of
  the tidal thames} (\bibinfo{year}{2021}).

\bibitem{beck2007statistics}
\bibinfo{author}{Beck, C.}
\newblock \bibinfo{title}{Statistics of three-dimensional lagrangian
  turbulence}.
\newblock \emph{\bibinfo{journal}{Physical Review Letters}}
  \textbf{\bibinfo{volume}{98}}, \bibinfo{pages}{064502}
  (\bibinfo{year}{2007}).

\bibitem{PhysRevE.72.056133}
\bibinfo{author}{Beck, C.}, \bibinfo{author}{Cohen, E. G.~D.} \&
  \bibinfo{author}{Swinney, H.~L.}
\newblock \bibinfo{title}{From time series to superstatistics}.
\newblock \emph{\bibinfo{journal}{Phys. Rev. E}} \textbf{\bibinfo{volume}{72}},
  \bibinfo{pages}{056133} (\bibinfo{year}{2005}).
\newblock \urlprefix\url{https://link.aps.org/doi/10.1103/PhysRevE.72.056133}.

\bibitem{BECK2003267}
\bibinfo{author}{Beck, C.} \& \bibinfo{author}{Cohen, E.}
\newblock \bibinfo{title}{Superstatistics}.
\newblock \emph{\bibinfo{journal}{Physica A: Statistical Mechanics and its
  Applications}} \textbf{\bibinfo{volume}{322}}, \bibinfo{pages}{267--275}
  (\bibinfo{year}{2003}).
\newblock
  \urlprefix\url{https://www.sciencedirect.com/science/article/pii/S0378437103000190}.

\bibitem{beck2009superstatistics}
\bibinfo{author}{Beck, C.}
\newblock \bibinfo{title}{Superstatistics in high-energy physics: application
  to cosmic ray energy spectra and e+ e-annihilation}.
\newblock \emph{\bibinfo{journal}{The European Physical Journal A}}
  \textbf{\bibinfo{volume}{40}}, \bibinfo{pages}{267} (\bibinfo{year}{2009}).

\bibitem{Sevilla2019}
\bibinfo{author}{Sevilla, F.~J.}, \bibinfo{author}{Arzola, A.~V.} \&
  \bibinfo{author}{Cital, E.~P.}
\newblock \bibinfo{title}{Stationary superstatistics distributions of trapped
  run-and-tumble particles}.
\newblock \emph{\bibinfo{journal}{Physical Review E}}
  \textbf{\bibinfo{volume}{99}} (\bibinfo{year}{2019}).
\newblock
  \urlprefix\url{https://www.scopus.com/inward/record.uri?eid=2-s2.0-85061244491&doi=10.1103%2fPhysRevE.99.012145&partnerID=40&md5=ab12ff85a1b7f937ce5583bd82902f52}.

\bibitem{Ayala2020}
\bibinfo{author}{Ayala, A.}, \bibinfo{author}{Hernández-Ortiz, S.},
  \bibinfo{author}{Hernández, L.}, \bibinfo{author}{Knapp-Pérez, V.} \&
  \bibinfo{author}{Zamora, R.}
\newblock \bibinfo{title}{Fluctuating temperature and baryon chemical potential
  in heavy-ion collisions and the position of the critical end point in the
  effective qcd phase diagram}.
\newblock \emph{\bibinfo{journal}{Physical Review D}}
  \textbf{\bibinfo{volume}{101}} (\bibinfo{year}{2020}).
\newblock
  \urlprefix\url{https://www.scopus.com/inward/record.uri?eid=2-s2.0-85084537884&doi=10.1103%2fPhysRevD.101.074023&partnerID=40&md5=e66eaa4d3ff146c92a46a410f0dda9a2}.

\bibitem{Cheraghalizadeh2021}
\bibinfo{author}{Cheraghalizadeh, J.}, \bibinfo{author}{Seifi, M.},
  \bibinfo{author}{Ebadi, Z.}, \bibinfo{author}{Mohammadzadeh, H.} \&
  \bibinfo{author}{Najafi, M.}
\newblock \bibinfo{title}{Superstatistical two-temperature ising model}.
\newblock \emph{\bibinfo{journal}{Physical Review E}}
  \textbf{\bibinfo{volume}{103}} (\bibinfo{year}{2021}).
\newblock
  \urlprefix\url{https://www.scopus.com/inward/record.uri?eid=2-s2.0-85102925395&doi=10.1103%2fPhysRevE.103.032104&partnerID=40&md5=70f36509adae22cbb805749fbac2729c}.

\bibitem{Yalcin2018}
\bibinfo{author}{Yalcin, G.~C.} \& \bibinfo{author}{Beck, C.}
\newblock \bibinfo{title}{Generalized statistical mechanics of cosmic rays:
  Application to positron-electron spectral indices}.
\newblock \emph{\bibinfo{journal}{Scientific Reports}}
  \textbf{\bibinfo{volume}{8}} (\bibinfo{year}{2018}).
\newblock
  \urlprefix\url{https://www.scopus.com/inward/record.uri?eid=2-s2.0-85041282876&doi=10.1038%2fs41598-018-20036-6&partnerID=40&md5=2fd796a5d12897eca4a9184680378fc6}.

\bibitem{Smolla_2020}
\bibinfo{author}{Smolla, M.}, \bibinfo{author}{Schäfer, B.},
  \bibinfo{author}{Lesch, H.} \& \bibinfo{author}{Beck, C.}
\newblock \bibinfo{title}{Universal properties of primary and secondary cosmic
  ray energy spectra}.
\newblock \emph{\bibinfo{journal}{New Journal of Physics}}
  \textbf{\bibinfo{volume}{22}}, \bibinfo{pages}{093002}
  (\bibinfo{year}{2020}).
\newblock \urlprefix\url{https://dx.doi.org/10.1088/1367-2630/abaa03}.

\bibitem{livadiotis2017kappa}
\bibinfo{author}{Livadiotis, G.}
\newblock \emph{\bibinfo{title}{Kappa distributions: Theory and applications in
  plasmas}} (\bibinfo{publisher}{Elsevier}, \bibinfo{year}{2017}).

\bibitem{Davis2019}
\bibinfo{author}{Davis, S.} \emph{et~al.}
\newblock \bibinfo{title}{Single-particle velocity distributions of
  collisionless, steady-state plasmas must follow superstatistics}.
\newblock \emph{\bibinfo{journal}{Physical Review E}}
  \textbf{\bibinfo{volume}{100}} (\bibinfo{year}{2019}).
\newblock
  \urlprefix\url{https://www.scopus.com/inward/record.uri?eid=2-s2.0-85072119582&doi=10.1103%2fPhysRevE.100.023205&partnerID=40&md5=50a421caf8116de6d350112f7c6aff58}.

\bibitem{PhysRevLett.118.143401}
\bibinfo{author}{Rouse, I.} \& \bibinfo{author}{Willitsch, S.}
\newblock \bibinfo{title}{Superstatistical energy distributions of an ion in an
  ultracold buffer gas}.
\newblock \emph{\bibinfo{journal}{Phys. Rev. Lett.}}
  \textbf{\bibinfo{volume}{118}}, \bibinfo{pages}{143401}
  (\bibinfo{year}{2017}).
\newblock
  \urlprefix\url{https://link.aps.org/doi/10.1103/PhysRevLett.118.143401}.

\bibitem{PhysRevX.7.021002}
\bibinfo{author}{Chechkin, A.~V.}, \bibinfo{author}{Seno, F.},
  \bibinfo{author}{Metzler, R.} \& \bibinfo{author}{Sokolov, I.~M.}
\newblock \bibinfo{title}{Brownian yet non-gaussian diffusion: From
  superstatistics to subordination of diffusing diffusivities}.
\newblock \emph{\bibinfo{journal}{Phys. Rev. X}} \textbf{\bibinfo{volume}{7}},
  \bibinfo{pages}{021002} (\bibinfo{year}{2017}).
\newblock \urlprefix\url{https://link.aps.org/doi/10.1103/PhysRevX.7.021002}.

\bibitem{Itto2021}
\bibinfo{author}{Itto, Y.} \& \bibinfo{author}{Beck, C.}
\newblock \bibinfo{title}{Superstatistical modelling of protein diffusion
  dynamics in bacteria}.
\newblock \emph{\bibinfo{journal}{Journal of the Royal Society Interface}}
  \textbf{\bibinfo{volume}{18}} (\bibinfo{year}{2021}).
\newblock
  \urlprefix\url{https://www.scopus.com/inward/record.uri?eid=2-s2.0-85102331224&doi=10.1098%2frsif.2020.0927&partnerID=40&md5=5253497da8c4ffb25592c212dd1f9e18}.

\bibitem{schafer2018non}
\bibinfo{author}{Sch{\"a}fer, B.}, \bibinfo{author}{Beck, C.},
  \bibinfo{author}{Aihara, K.}, \bibinfo{author}{Witthaut, D.} \&
  \bibinfo{author}{Timme, M.}
\newblock \bibinfo{title}{Non-gaussian power grid frequency fluctuations
  characterized by l{\'e}vy-stable laws and superstatistics}.
\newblock \emph{\bibinfo{journal}{Nature Energy}} \textbf{\bibinfo{volume}{3}},
  \bibinfo{pages}{119--126} (\bibinfo{year}{2018}).

\bibitem{Bogachev20171}
\bibinfo{author}{Bogachev, V.} \& \bibinfo{author}{Smolyanov, O.}
\newblock \bibinfo{title}{Introduction to the theory of topological vector
  spaces}.
\newblock \emph{\bibinfo{journal}{Springer Monographs in Mathematics}}
  \bibinfo{pages}{1 – 100} (\bibinfo{year}{2017}).
\newblock
  \urlprefix\url{https://www.scopus.com/inward/record.uri?eid=2-s2.0-85020414563&doi=10.1007%2f978-3-319-57117-1_1&partnerID=40&md5=6ed9ab90b3ebf18f247461632b89d8f2}.

\bibitem{GIDEA2018820}
\bibinfo{author}{Gidea, M.} \& \bibinfo{author}{Katz, Y.}
\newblock \bibinfo{title}{Topological data analysis of financial time series:
  Landscapes of crashes}.
\newblock \emph{\bibinfo{journal}{Physica A: Statistical Mechanics and its
  Applications}} \textbf{\bibinfo{volume}{491}}, \bibinfo{pages}{820--834}
  (\bibinfo{year}{2018}).
\newblock
  \urlprefix\url{https://www.sciencedirect.com/science/article/pii/S0378437117309202}.

\bibitem{UCHIYAMA2019120930}
\bibinfo{author}{Uchiyama, Y.} \& \bibinfo{author}{Kadoya, T.}
\newblock \bibinfo{title}{Superstatistics with cut-off tails for financial time
  series}.
\newblock \emph{\bibinfo{journal}{Physica A: Statistical Mechanics and its
  Applications}} \textbf{\bibinfo{volume}{526}}, \bibinfo{pages}{120930}
  (\bibinfo{year}{2019}).
\newblock
  \urlprefix\url{https://www.sciencedirect.com/science/article/pii/S0378437119305345}.

\bibitem{BRIGGS2007498}
\bibinfo{author}{Briggs, K.} \& \bibinfo{author}{Beck, C.}
\newblock \bibinfo{title}{Modelling train delays with q-exponential functions}.
\newblock \emph{\bibinfo{journal}{Physica A: Statistical Mechanics and its
  Applications}} \textbf{\bibinfo{volume}{378}}, \bibinfo{pages}{498--504}
  (\bibinfo{year}{2007}).
\newblock
  \urlprefix\url{https://www.sciencedirect.com/science/article/pii/S0378437106013227}.

\bibitem{mitsokapas2021statistical}
\bibinfo{author}{Mitsokapas, E.}, \bibinfo{author}{Sch{\"a}fer, B.},
  \bibinfo{author}{Harris, R.~J.} \& \bibinfo{author}{Beck, C.}
\newblock \bibinfo{title}{Statistical characterization of airplane delays}.
\newblock \emph{\bibinfo{journal}{Scientific Reports}}
  \textbf{\bibinfo{volume}{11}}, \bibinfo{pages}{7855} (\bibinfo{year}{2021}).

\bibitem{Weber20191}
\bibinfo{author}{Weber, E.~J.}
\newblock \bibinfo{title}{Highlights from this issue}.
\newblock \emph{\bibinfo{journal}{Emergency Medicine Journal}}
  \textbf{\bibinfo{volume}{36}}, \bibinfo{pages}{1} (\bibinfo{year}{2019}).
\newblock
  \urlprefix\url{https://www.scopus.com/inward/record.uri?eid=2-s2.0-85059878468&doi=10.1136%2femermed-2018-208324&partnerID=40&md5=b9fa02bf967d36548437590ab4302b1c}.

\bibitem{Williams2020}
\bibinfo{author}{Williams, G.}, \bibinfo{author}{Schäfer, B.} \&
  \bibinfo{author}{Beck, C.}
\newblock \bibinfo{title}{Superstatistical approach to air pollution
  statistics}.
\newblock \emph{\bibinfo{journal}{Physical Review Research}}
  \textbf{\bibinfo{volume}{2}} (\bibinfo{year}{2020}).
\newblock
  \urlprefix\url{https://www.scopus.com/inward/record.uri?eid=2-s2.0-85090293965&doi=10.1103%2fPhysRevResearch.2.013019&partnerID=40&md5=232925cecda24a3d0d067b42143f337e}.

\bibitem{he2022spatial}
\bibinfo{author}{He, H.}, \bibinfo{author}{Sch{\"a}fer, B.} \&
  \bibinfo{author}{Beck, C.}
\newblock \bibinfo{title}{Spatial heterogeneity of air pollution statistics in
  europe}.
\newblock \emph{\bibinfo{journal}{Scientific Reports}}
  \textbf{\bibinfo{volume}{12}}, \bibinfo{pages}{12215} (\bibinfo{year}{2022}).

\bibitem{de2018superstatistical}
\bibinfo{author}{De~Michele, C.} \& \bibinfo{author}{Avanzi, F.}
\newblock \bibinfo{title}{Superstatistical distribution of daily precipitation
  extremes: a worldwide assessment. sci. rep. 8 (1), 1--11}
  (\bibinfo{year}{2018}).

\bibitem{npg-30-515-2023}
\bibinfo{author}{Flora, S.}, \bibinfo{author}{Ursella, L.} \&
  \bibinfo{author}{Wirth, A.}
\newblock \bibinfo{title}{Superstatistical analysis of sea surface currents in
  the gulf of trieste, measured by high-frequency radar, and its relation to
  wind regimes using the maximum-entropy principle}.
\newblock \emph{\bibinfo{journal}{Nonlinear Processes in Geophysics}}
  \textbf{\bibinfo{volume}{30}}, \bibinfo{pages}{515--525}
  (\bibinfo{year}{2023}).
\newblock \urlprefix\url{https://npg.copernicus.org/articles/30/515/2023/}.

\bibitem{Sardeli_2023}
\bibinfo{author}{Sardeli, E.} \emph{et~al.}
\newblock \bibinfo{title}{Complexity of recent earthquake swarms in greece in
  terms of non-extensive statistical physics}.
\newblock \emph{\bibinfo{journal}{Entropy}} \textbf{\bibinfo{volume}{25}},
  \bibinfo{pages}{667} (\bibinfo{year}{2023}).
\newblock \urlprefix\url{http://dx.doi.org/10.3390/e25040667}.

\bibitem{agarap2018deep}
\bibinfo{author}{Agarap, A.~F.}
\newblock \bibinfo{title}{Deep learning using rectified linear units (relu)}.
\newblock \emph{\bibinfo{journal}{arXiv preprint arXiv:1803.08375}}
  (\bibinfo{year}{2018}).

\bibitem{SCHAFER2021102881}
\bibinfo{author}{Schäfer, B.}, \bibinfo{author}{Heppell, C.~M.},
  \bibinfo{author}{Rhys, H.} \& \bibinfo{author}{Beck, C.}
\newblock \bibinfo{title}{Fluctuations of water quality time series in rivers
  follow superstatistics}.
\newblock \emph{\bibinfo{journal}{iScience}} \textbf{\bibinfo{volume}{24}},
  \bibinfo{pages}{102881} (\bibinfo{year}{2021}).
\newblock
  \urlprefix\url{https://www.sciencedirect.com/science/article/pii/S258900422100849X}.

\bibitem{schafer2022machine}
\bibinfo{author}{Sch{\"a}fer, B.} \emph{et~al.}
\newblock \bibinfo{title}{Machine learning approach towards explaining water
  quality dynamics in an urbanised river}.
\newblock \emph{\bibinfo{journal}{Scientific Reports}}
  \textbf{\bibinfo{volume}{12}}, \bibinfo{pages}{12346} (\bibinfo{year}{2022}).

\bibitem{10.1007/978-3-031-43169-2_21}
\bibinfo{author}{Citakoglu, H.}, \bibinfo{author}{Ozeren, Y.} \&
  \bibinfo{author}{Gemici, B.~T.}
\newblock \bibinfo{title}{Prediction machine learning methods for dissolved
  oxygen value of the sakarya basin in turkey}.
\newblock In \bibinfo{editor}{Chenchouni, H.} \emph{et~al.} (eds.)
  \emph{\bibinfo{booktitle}{Recent Research on Hydrogeology, Geoecology and
  Atmospheric Sciences}}, \bibinfo{pages}{95--98} (\bibinfo{publisher}{Springer
  Nature Switzerland}, \bibinfo{address}{Cham}, \bibinfo{year}{2023}).

\bibitem{mohamed2018suspended}
\bibinfo{author}{Mohamed, I.} \& \bibinfo{author}{Shah, I.}
\newblock \bibinfo{title}{Suspended sediment concentration modeling using
  conventional and machine learning approaches in the thames river, london
  ontario}.
\newblock \emph{\bibinfo{journal}{Journal of Water Management Modeling}}
  (\bibinfo{year}{2018}).

\bibitem{ASADOLLAH2021104599}
\bibinfo{author}{Asadollah, S. B. H.~S.}, \bibinfo{author}{Sharafati, A.},
  \bibinfo{author}{Motta, D.} \& \bibinfo{author}{Yaseen, Z.~M.}
\newblock \bibinfo{title}{River water quality index prediction and uncertainty
  analysis: A comparative study of machine learning models}.
\newblock \emph{\bibinfo{journal}{Journal of Environmental Chemical
  Engineering}} \textbf{\bibinfo{volume}{9}}, \bibinfo{pages}{104599}
  (\bibinfo{year}{2021}).
\newblock
  \urlprefix\url{https://www.sciencedirect.com/science/article/pii/S2213343720309489}.

\bibitem{doi:10.1021/acs.est.0c06783}
\bibinfo{author}{Zhi, W.} \emph{et~al.}
\newblock \bibinfo{title}{From hydrometeorology to river water quality: Can a
  deep learning model predict dissolved oxygen at the continental scale?}
\newblock \emph{\bibinfo{journal}{Environmental Science \& Technology}}
  \textbf{\bibinfo{volume}{55}}, \bibinfo{pages}{2357--2368}
  (\bibinfo{year}{2021}).
\newblock \urlprefix\url{https://doi.org/10.1021/acs.est.0c06783}.
\newblock \bibinfo{note}{PMID: 33533608},
  \eprint{https://doi.org/10.1021/acs.est.0c06783}.

\bibitem{castrillo2020estimation}
\bibinfo{author}{Castrillo, M.} \& \bibinfo{author}{Garc{\'\i}a, {\'A}.~L.}
\newblock \bibinfo{title}{Estimation of high frequency nutrient concentrations
  from water quality surrogates using machine learning methods}.
\newblock \emph{\bibinfo{journal}{Water research}}
  \textbf{\bibinfo{volume}{172}}, \bibinfo{pages}{115490}
  (\bibinfo{year}{2020}).

\bibitem{SAINZPARDODIAZ2023120726}
\bibinfo{author}{{Sáinz-Pardo Díaz}, J.}, \bibinfo{author}{Castrillo, M.} \&
  \bibinfo{author}{Álvaro {López García}}.
\newblock \bibinfo{title}{Deep learning based soft-sensor for continuous
  chlorophyll estimation on decentralized data}.
\newblock \emph{\bibinfo{journal}{Water Research}}
  \textbf{\bibinfo{volume}{246}}, \bibinfo{pages}{120726}
  (\bibinfo{year}{2023}).
\newblock
  \urlprefix\url{https://www.sciencedirect.com/science/article/pii/S0043135423011661}.

\bibitem{https://doi.org/10.1002/lom3.10468}
\bibinfo{author}{Maguire, T.~J.}, \bibinfo{author}{Dominato, K.~R.},
  \bibinfo{author}{Weidman, R.~P.} \& \bibinfo{author}{Mundle, S. O.~C.}
\newblock \bibinfo{title}{Ultraviolet-visual spectroscopy estimation of nitrate
  concentrations in surface waters via machine learning}.
\newblock \emph{\bibinfo{journal}{Limnology and Oceanography: Methods}}
  \textbf{\bibinfo{volume}{20}}, \bibinfo{pages}{26--33}
  (\bibinfo{year}{2022}).
\newblock
  \urlprefix\url{https://aslopubs.onlinelibrary.wiley.com/doi/abs/10.1002/lom3.10468}.
\newblock
  \eprint{https://aslopubs.onlinelibrary.wiley.com/doi/pdf/10.1002/lom3.10468}.

\bibitem{zounemat2021comprehensive}
\bibinfo{author}{Zounemat-Kermani, M.}, \bibinfo{author}{Mahdavi-Meymand, A.}
  \& \bibinfo{author}{Hinkelmann, R.}
\newblock \bibinfo{title}{A comprehensive survey on conventional and modern
  neural networks: application to river flow forecasting}.
\newblock \emph{\bibinfo{journal}{Earth Science Informatics}}
  \textbf{\bibinfo{volume}{14}}, \bibinfo{pages}{893--911}
  (\bibinfo{year}{2021}).

\bibitem{vaswani2017attention}
\bibinfo{author}{Vaswani, A.} \emph{et~al.}
\newblock \bibinfo{title}{Attention is all you need}.
\newblock \emph{\bibinfo{journal}{Advances in neural information processing
  systems}} \textbf{\bibinfo{volume}{30}} (\bibinfo{year}{2017}).

\bibitem{nair2023temporal}
\bibinfo{author}{Nair, J.~P.} \& \bibinfo{author}{Vijaya, M.}
\newblock \bibinfo{title}{Temporal fusion transformer: A deep learning approach
  for modeling and forecasting river water quality index}.
\newblock \emph{\bibinfo{journal}{International Journal of Intelligent Systems
  and Applications in Engineering}} \textbf{\bibinfo{volume}{11}},
  \bibinfo{pages}{277--293} (\bibinfo{year}{2023}).

\bibitem{beck2004superstatistics}
\bibinfo{author}{Beck, C.}
\newblock \bibinfo{title}{Superstatistics in hydrodynamic turbulence}.
\newblock \emph{\bibinfo{journal}{Physica D: Nonlinear Phenomena}}
  \textbf{\bibinfo{volume}{193}}, \bibinfo{pages}{195--207}
  (\bibinfo{year}{2004}).

\bibitem{Meteor}
\bibinfo{author}{{Meteor Communications}}.
\newblock \bibinfo{title}{Water quality monitoring systems \& services}.
\newblock
  \bibinfo{howpublished}{\url{https://meteorcommunications.co.uk/water-quality-monitoring/}}.
\newblock \bibinfo{note}{Accessed: [2022]}.

\bibitem{Folium}
\bibinfo{author}{Contributors, F.}
\newblock \bibinfo{title}{Folium: Python data, leaflet.js maps}.
\newblock
  \bibinfo{howpublished}{\url{https://github.com/python-visualization/folium}}.

\bibitem{OpenStreetMap}
\bibinfo{author}{Contributors, O.}
\newblock \bibinfo{title}{Openstreetmap}.
\newblock \bibinfo{howpublished}{\url{https://www.openstreetmap.org}}
  (\bibinfo{year}{2023}).

\bibitem{tsallis1988possible}
\bibinfo{author}{Tsallis, C.}
\newblock \bibinfo{title}{Possible generalization of boltzmann-gibbs
  statistics}.
\newblock \emph{\bibinfo{journal}{Journal of Statistical Physics}}
  \textbf{\bibinfo{volume}{52}}, \bibinfo{pages}{479--487}
  (\bibinfo{year}{1988}).

\bibitem{beck2001dynamical}
\bibinfo{author}{Beck, C.}
\newblock \bibinfo{title}{Dynamical foundations of nonextensive statistical
  mechanics}.
\newblock \emph{\bibinfo{journal}{Physical Review Letters}}
  \textbf{\bibinfo{volume}{87}}, \bibinfo{pages}{180601}
  (\bibinfo{year}{2001}).

\bibitem{ke2017lightgbm}
\bibinfo{author}{Ke, G.} \emph{et~al.}
\newblock \bibinfo{title}{Lightgbm: A highly efficient gradient boosting
  decision tree}.
\newblock \emph{\bibinfo{journal}{Advances in neural information processing
  systems}} \textbf{\bibinfo{volume}{30}} (\bibinfo{year}{2017}).

\bibitem{lundberg2017unified}
\bibinfo{author}{Lundberg, S.~M.} \& \bibinfo{author}{Lee, S.-I.}
\newblock \bibinfo{title}{A unified approach to interpreting model
  predictions}.
\newblock \emph{\bibinfo{journal}{Advances in neural information processing
  systems}} \textbf{\bibinfo{volume}{30}} (\bibinfo{year}{2017}).

\bibitem{molnar2020interpretable}
\bibinfo{author}{Molnar, C.}
\newblock \emph{\bibinfo{title}{Interpretable Machine Learning}}
  (\bibinfo{publisher}{Lulu. com}, \bibinfo{year}{2020}).

\bibitem{zhou2021informer}
\bibinfo{author}{Zhou, H.} \emph{et~al.}
\newblock \bibinfo{title}{Informer: Beyond efficient transformer for long
  sequence time-series forecasting}.
\newblock In \emph{\bibinfo{booktitle}{Proceedings of the AAAI conference on
  artificial intelligence}}, vol.~\bibinfo{volume}{35},
  \bibinfo{pages}{11106--11115} (\bibinfo{year}{2021}).

\bibitem{huang2017densely}
\bibinfo{author}{Huang, G.}, \bibinfo{author}{Liu, Z.}, \bibinfo{author}{Van
  Der~Maaten, L.} \& \bibinfo{author}{Weinberger, K.~Q.}
\newblock \bibinfo{title}{Densely connected convolutional networks}.
\newblock In \emph{\bibinfo{booktitle}{Proceedings of the IEEE conference on
  computer vision and pattern recognition}}, \bibinfo{pages}{4700--4708}
  (\bibinfo{year}{2017}).

\bibitem{o2015introduction}
\bibinfo{author}{O'Shea, K.} \& \bibinfo{author}{Nash, R.}
\newblock \bibinfo{title}{An introduction to convolutional neural networks}.
\newblock \emph{\bibinfo{journal}{arXiv preprint arXiv:1511.08458}}
  (\bibinfo{year}{2015}).

\bibitem{hochreiter1997long}
\bibinfo{author}{Hochreiter, S.} \& \bibinfo{author}{Schmidhuber, J.}
\newblock \bibinfo{title}{Long short-term memory}.
\newblock \emph{\bibinfo{journal}{Neural computation}}
  \textbf{\bibinfo{volume}{9}}, \bibinfo{pages}{1735--1780}
  (\bibinfo{year}{1997}).

\bibitem{lim2021temporal}
\bibinfo{author}{Lim, B.}, \bibinfo{author}{Ar{\i}k, S.~{\"O}.},
  \bibinfo{author}{Loeff, N.} \& \bibinfo{author}{Pfister, T.}
\newblock \bibinfo{title}{Temporal fusion transformers for interpretable
  multi-horizon time series forecasting}.
\newblock \emph{\bibinfo{journal}{International Journal of Forecasting}}
  \textbf{\bibinfo{volume}{37}}, \bibinfo{pages}{1748--1764}
  (\bibinfo{year}{2021}).

\bibitem{Challu_Olivares_Oreshkin_Garza}
\bibinfo{author}{Challu, C.} \emph{et~al.}
\newblock \bibinfo{title}{Nhits: Neural hierarchical interpolation for time
  series forecasting}.
\newblock \emph{\bibinfo{journal}{Proceedings of the AAAI Conference on
  Artificial Intelligence}} \textbf{\bibinfo{volume}{37}},
  \bibinfo{pages}{6989--6997} (\bibinfo{year}{2023}).
\newblock
  \urlprefix\url{https://ojs.aaai.org/index.php/AAAI/article/view/25854}.

\bibitem{NEURIPS2021_c68bd905}
\bibinfo{author}{Tang, B.} \& \bibinfo{author}{Matteson, D.~S.}
\newblock \bibinfo{title}{Probabilistic transformer for time series analysis}.
\newblock In \bibinfo{editor}{Ranzato, M.}, \bibinfo{editor}{Beygelzimer, A.},
  \bibinfo{editor}{Dauphin, Y.}, \bibinfo{editor}{Liang, P.} \&
  \bibinfo{editor}{Vaughan, J.~W.} (eds.) \emph{\bibinfo{booktitle}{Advances in
  Neural Information Processing Systems}}, vol.~\bibinfo{volume}{34},
  \bibinfo{pages}{23592--23608} (\bibinfo{publisher}{Curran Associates, Inc.},
  \bibinfo{year}{2021}).
\newblock
  \urlprefix\url{https://proceedings.neurips.cc/paper_files/paper/2021/file/c68bd9055776bf38d8fc43c0ed283678-Paper.pdf}.

\bibitem{pmlr-v119-duan20a}
\bibinfo{author}{Duan, T.} \emph{et~al.}
\newblock \bibinfo{title}{{NGB}oost: Natural gradient boosting for
  probabilistic prediction}.
\newblock In \bibinfo{editor}{III, H.~D.} \& \bibinfo{editor}{Singh, A.} (eds.)
  \emph{\bibinfo{booktitle}{Proceedings of the 37th International Conference on
  Machine Learning}}, vol. \bibinfo{volume}{119} of
  \emph{\bibinfo{series}{Proceedings of Machine Learning Research}},
  \bibinfo{pages}{2690--2700} (\bibinfo{publisher}{PMLR},
  \bibinfo{year}{2020}).
\newblock \urlprefix\url{https://proceedings.mlr.press/v119/duan20a.html}.

\bibitem{midas_daily}
\bibinfo{author}{{Met Office}}.
\newblock \bibinfo{title}{Midas: Uk daily rainfall data}.
\newblock \bibinfo{howpublished}{NCAS British Atmospheric Data Centre}
  (\bibinfo{year}{2006}).
\newblock
  \urlprefix\url{https://catalogue.ceda.ac.uk/uuid/c732716511d3442f05cdeccbe99b8f90}.
\newblock \bibinfo{note}{[Date of citation]}.

\bibitem{midas_hourly}
\bibinfo{author}{{Met Office}}.
\newblock \bibinfo{title}{Midas uk hourly rainfall data}.
\newblock \bibinfo{howpublished}{NCAS British Atmospheric Data Centre}
  (\bibinfo{year}{2006}).
\newblock
  \urlprefix\url{https://catalogue.ceda.ac.uk/uuid/bbd6916225e7475514e17fdbf11141c1}.
\newblock \bibinfo{note}{[Date of citation]}.

\bibitem{wetzel2001limnology}
\bibinfo{author}{Wetzel, R.~G.}
\newblock \emph{\bibinfo{title}{Limnology: Lake and River Ecosystems}}
  (\bibinfo{publisher}{Gulf Professional Publishing}, \bibinfo{year}{2001}).
\newblock \bibinfo{note}{Chapter 9, p.152}.

\bibitem{Murugesu2023Thames}
\bibinfo{author}{Murugesu, J.~A.}
\newblock \bibinfo{title}{River thames was pumped full of oxygen in 2022 to
  prevent fish deaths}.
\newblock \emph{\bibinfo{journal}{New Scientist}}  (\bibinfo{year}{2023}).
\newblock
  \urlprefix\url{https://www.newscientist.com/article/2379513-river-thames-was-pumped-full-of-oxygen-in-2022-to-prevent-fish-deaths/}.

\bibitem{statsmodels2023}
\bibinfo{title}{{statsmodels.tsa.seasonal}}.
\newblock
  \bibinfo{howpublished}{\url{https://www.statsmodels.org/stable/generated/statsmodels.tsa.seasonal.seasonal_decompose.html}}
  (\bibinfo{year}{2023}).
\newblock \bibinfo{note}{Accessed: {2023}}.

\bibitem{pyemd}
\bibinfo{author}{Laszuk, D.}
\newblock \bibinfo{title}{Python implementation of empirical mode decomposition
  algorithm}.
\newblock \bibinfo{howpublished}{\url{https://github.com/laszukdawid/PyEMD}}
  (\bibinfo{year}{2017}).
\newblock \bibinfo{note}{Accessed: [2023]}.

\bibitem{akiba2019optuna}
\bibinfo{author}{Akiba, T.}, \bibinfo{author}{Sano, S.},
  \bibinfo{author}{Yanase, T.}, \bibinfo{author}{Ohta, T.} \&
  \bibinfo{author}{Koyama, M.}
\newblock \bibinfo{title}{Optuna: A next-generation hyperparameter optimization
  framework}.
\newblock In \emph{\bibinfo{booktitle}{Proceedings of the 25th ACM SIGKDD
  international conference on knowledge discovery \& data mining}},
  \bibinfo{pages}{2623--2631} (\bibinfo{year}{2019}).

\bibitem{726791}
\bibinfo{author}{Lecun, Y.}, \bibinfo{author}{Bottou, L.},
  \bibinfo{author}{Bengio, Y.} \& \bibinfo{author}{Haffner, P.}
\newblock \bibinfo{title}{Gradient-based learning applied to document
  recognition}.
\newblock \emph{\bibinfo{journal}{Proceedings of the IEEE}}
  \textbf{\bibinfo{volume}{86}}, \bibinfo{pages}{2278--2324}
  (\bibinfo{year}{1998}).

\bibitem{kingma2014adam}
\bibinfo{author}{Kingma, D.~P.} \& \bibinfo{author}{Ba, J.}
\newblock \bibinfo{title}{Adam: A method for stochastic optimization}.
\newblock \emph{\bibinfo{journal}{arXiv preprint arXiv:1412.6980}}
  (\bibinfo{year}{2014}).

\end{thebibliography}

\subsection*{Acknowledgments}
We thank Sebastian Pütz for helpful discussions on the machine learning aspects. Moreover, we thank Pippa Tucker from the Environmental Agency for providing us with some of the data. We gratefully acknowledge funding from the Helmholtz Association and the Networking Fund through Helmholtz AI and under grant no. VH-NG-1727. This research was also funded by a QMUL Research England impact fund with the title ``Machine Learning Algorithm for Water Quality Prediction". We are also grateful to Oracle for awarding the Cloud Starter Award, which provided the computational resources needed for this research. 

\subsection*{Author contributions}
All authors conceived the project. K.H., H.H., and C.B. curated the data. H.H., T.B., and B.S. processed the data, performed the data analysis, and produced the plots. H.H., T.B., and C.B. wrote the original draft. All authors interpreted the results and revised the manuscript.

\subsection*{Competing interests}
The authors declare no competing interests.

\subsection*{Additional information}
\textbf{Correspondence} and requests for materials should be addressed to H.H.

\end{document}